\documentclass[10pt,journal,compsoc]{IEEEtran}

\usepackage{cite}

\ifCLASSINFOpdf
   \usepackage[pdftex]{graphicx}
  
\else
  
\fi

\usepackage{amsmath}
\usepackage{amssymb}
\usepackage{amstext}
\usepackage[T1]{fontenc}

\usepackage{tabulary,xfrac,enumitem}
\usepackage[dvipsnames]{xcolor}
\usepackage[ruled,noresetcount]{algorithm2e}
\usepackage[pagebackref=false,breaklinks=true,colorlinks,bookmarks=false]{hyperref}
\definecolor{citecolor}{RGB}{34,139,34}

\usepackage{array}

\usepackage{url}

\usepackage{booktabs}
\usepackage{multirow}
\usepackage{threeparttable}
\usepackage{color, colortbl}
\usepackage{changes}
\usepackage{subcaption}
\usepackage{wrapfig}
\usepackage{diagbox}
\usepackage{arydshln} 
\usepackage{makecell}

\usepackage{pifont}

\makeatletter
\newcommand{\thickhline}{%
    \noalign {\ifnum 0=`}\fi \hrule height 0.5pt
    \futurelet \reserved@a \@xhline
}
\makeatother
\definecolor{LightGray}{gray}{0.9}

\begin{document}

\title{From Zero to Detail: A Progressive Spectral Decoupling Paradigm for UHD Image Restoration with New Benchmark}
\newcommand{\zongxin}[1]{{#1}}
\newcommand{\new}[1]{{#1}}

\author{Chen Zhao, Member, IEEE, Yunzhe Xu, Zhizhou Chen, Enxuan Gu, Kai Zhang, 

Xiaoming Liu, Fellow, IEEE, Jian Yang, and Ying Tai$^{\dagger}$
\thanks{

Chen Zhao, Yunzhe Xu, Zhizhou Chen, Kai Zhang, Jian Yang, and Ying Tai are affiliated with the State Key Laboratory of Novel Software Technology,
Nanjing University, Nanjing 210023, China.

Xiaoming Liu is with the Department of Computer Science and Engineering,
 Michigan State University, East Lansing, MI 48824, U.S.A.

Enxuan Gu is with the School of Computer Science and Technology, Dalian University of Technology, Dalian 116023, China.
 
Chen Zhao and Yunzhe Xu contribute to this work equally.

Ying Tai is the corresponding author. E-mail: yingtai@nju.edu.cn.

}}

\markboth{IEEE TRANSACTIONS ON PATTERN ANALYSIS AND MACHINE INTELLIGENCE}%
{Shell \MakeLowercase{\textit{et al.}}: Bare Demo of IEEEtran.cls for Computer Society Journals}
\IEEEtitleabstractindextext{%
\begin{abstract}
Ultra-high-definition (UHD) image restoration poses unique challenges due to the high spatial resolution, diverse content, and fine-grained structures present in UHD images. To address these issues, we introduce a progressive spectral decomposition for the restoration process, decomposing it into three stages: zero-frequency \textbf{enhancement}, low-frequency \textbf{restoration}, and high-frequency \textbf{refinement}. Based on this formulation, we propose a novel framework, \textbf{ERR}, which integrates three  cooperative sub-networks: the zero-frequency enhancer (ZFE), the low-frequency restorer (LFR), and the high-frequency refiner (HFR). The ZFE incorporates global priors to learn holistic mappings, the LFR reconstructs the main content by focusing on coarse-scale information, and the HFR adopts our proposed frequency-windowed Kolmogorov-Arnold Network (FW-KAN) to recover fine textures and intricate details for high-fidelity restoration.
To further advance research in UHD image restoration, we also construct a large-scale, high-quality benchmark dataset, \textbf{LSUHDIR}, comprising 82{,}126 UHD images with diverse scenes and rich content. Our proposed methods demonstrate superior performance across a range of UHD image restoration tasks, and extensive ablation studies confirm the contribution and necessity of each module. Project page: \url{https://github.com/NJU-PCALab/ERR}.

\end{abstract}

}

\maketitle

\IEEEdisplaynontitleabstractindextext

\IEEEpeerreviewmaketitle

\IEEEraisesectionheading{\section{Introduction}\label{sec:introduction}}

With the  development of ultra-high-definition (UHD) imaging, its applications have expanded across a wide range of domains \cite{li2023embedding,yu2022towards,zheng2021ultra,deng2021multi,li2023uhdnerf,sun2024ultra}. Nonetheless, UHD images captured in challenging environments—such as under low illumination, adverse weather (e.g., rain, haze or snow), or noise interference—are often subject to significant quality degradation \cite{wang2024correlation,chen2024towards,li2023embedding}. This work focuses on addressing the problem of UHD image restoration (IR) under such adverse conditions.

The remarkable progress of deep learning has spurred the development of numerous learning-based approaches for addressing IR challenges, particularly those based on convolutional neural networks (CNNs) and Transformers \cite{liang2021swinir,zamir2022restormer,zhao2025learning,zhou2026more,zhao2025multi,dong2025mamba,zhao2024cycle,zhao2025spectral,xie2024addsr,hu2025exploiting,zhao2025ultrahr,zhao2026luve,DBLP:conf/iccv/LuLK23,zhao2026learning,zhou2024adapt}. Although recent methods demonstrate impressive performance, they are predominantly designed for general IR tasks, which limits their ability to achieve high-quality results for UHD IR. 
Current state-of-the-art (SOTA) methods primarily enhance network performance by introducing more complex architectures \cite{li2023efficient,zhang2024distilling,DBLP:conf/cvpr/0002FZL00Z24,zhou2024migc,lu2024mace,zhangrethinking,zhang2025gapt,zhang2025r,zhang2024vocapter,zhang2025u,zhou2025bidedpo}. However, due to the ultra-high resolution and pixel density of UHD images, these advanced methods struggle to perform effectively in UHD scenarios, constraining the potential applications of UHD imaging systems.


Recently, a number of methods have been specifically developed for UHD IR \cite{wang2024correlation,zou2024wave,wang2023ultra,zhou2023pyramid,zhou2024migc++,zhou20243dis,zhou2025dreamrenderer,zhou20253dis,xu2025contextgen,xu2026phyedit,DBLP:conf/mm/YuZZHZZ23,DBLP:conf/mm/XiaoLW24,liu2025dreamuhd}. For instance, LLFormer \cite{wang2023ultra} demonstrated impressive performance by exploiting the strong long-range modeling capabilities of Transformers. However, its high computational demands make it unsuitable for efficient full-resolution inference on edge devices. To address this, UHDFour \cite{li2023embedding} adopted an 8× downsampling strategy, enabling full-resolution inference of UHD images with reduced computational overhead. UHDFormer \cite{wang2024correlation} and UHDPromer \cite{wang2026neural} introduced a Transformer-based architecture that utilizes high-resolution features to guide the low-resolution restoration.
Although these approaches successfully reduce computational complexity through downsampling, the inherent \textit{downsampling–enhancement–upsampling learning paradigm} \cite{wang2025deep} inevitably results in the loss of critical information \cite{yu2024empowering}. Moreover, the nature of UHD images—characterized by \textit{ultra-high resolution, extensive content, and intricate structural details}—introduces substantial challenges for restoration, making it difficult for existing methods to achieve both efficiency and high-quality results.

\begin{figure*}[t]
	\centering
	\begin{subfigure}[t]{0.7\linewidth}
		\centering
		\includegraphics[width=\linewidth]{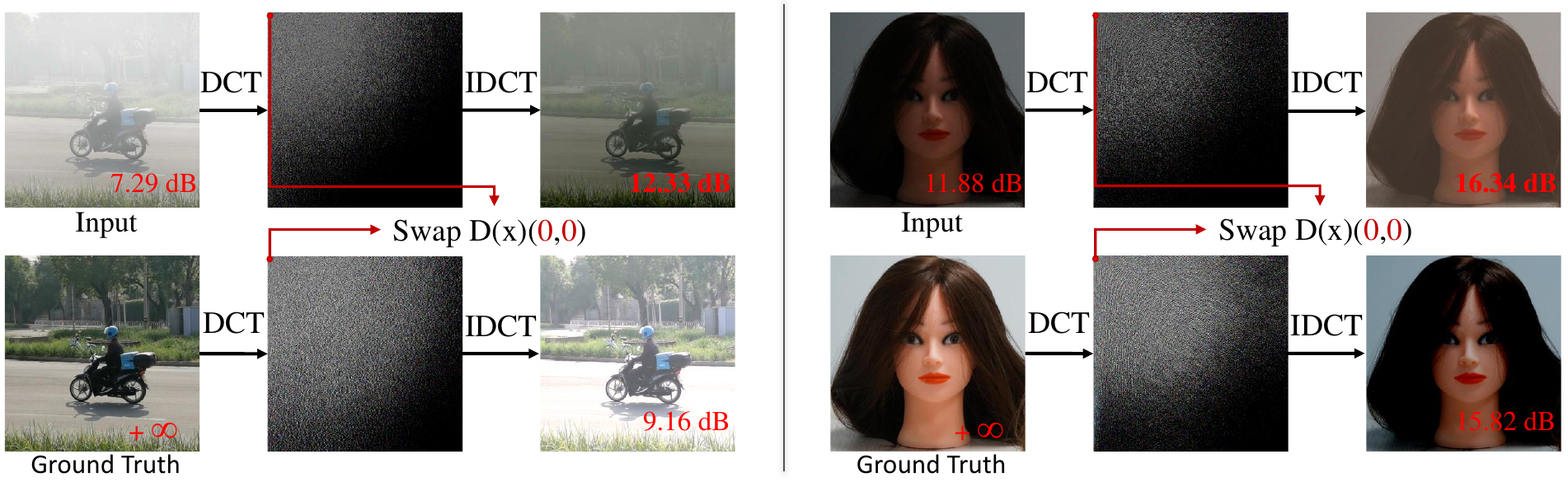}
		\caption{We exchange the information at the (0,0) position in the DCT spectrum. }
	\label{fig:2a}
	\end{subfigure}%
	
	\vskip 0.1cm  

	\begin{subfigure}[t]{0.85 \linewidth}
		\centering
		\includegraphics[width=\linewidth]{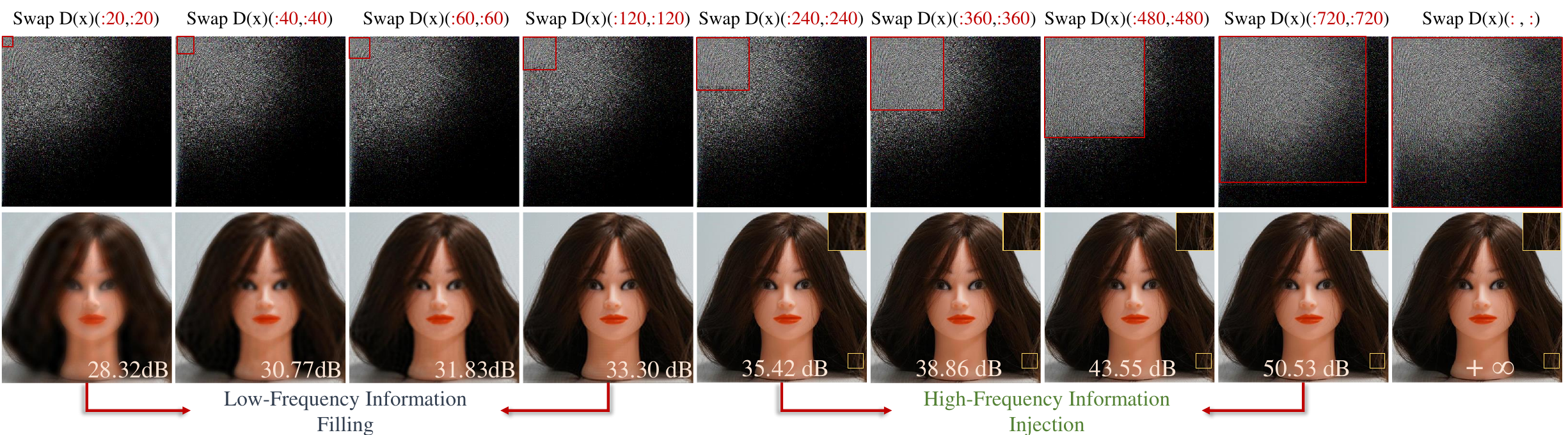}
		\caption{The low-frequency filling restores the coarse-grained 
content, while the high-frequency injection refines the fine-grained textures.
        }
	\label{fig:2b}
	\end{subfigure}
	
	\caption{Our core motivation. Based on the observations in (a) and (b), we decompose the complex UHD restoration problem into three progressive stages: zero-frequency enhancement, low-frequency restoration, and high-frequency refinement.
    }
    \vspace{-0.5em}
	\label{fig:2}
\end{figure*}

\begin{table*}[!t]
    \caption{\small{Comprehensive comparison with our conference version demonstrates that our IERR is \textbf{superior, smaller, and faster}.}}
     \vspace{-0.8em}
	\begin{center}
		\begin{tabular}{c|c  c c c  c  c   c   c   c}
			\hline
            PSNR/SSIM & UHD-LL   &  UHD-Haze  & UHD-Blur & 4K-Rain13k&Param(M)$\downarrow$&FLOPs$\downarrow$&Memory(MiB)$\downarrow$&latency(s)$\downarrow$ \\
			\hline 
            ERR  & 27.57/0.932 &  25.12/0.950 & 29.72/0.861 &   34.48/0.952 & 1.131& 307.52 & 9880 & 0.52 \\
            IERR  & \textbf{27.87}/\textbf{0.932} & \textbf{27.06}/\textbf{0.958}  &  \textbf{30.53}/\textbf{0.873} & \textbf{34.89}/\textbf{0.955}   & \textbf{0.503}& \textbf{288.81} & \textbf{9771} & \textbf{0.33} \\
			\hline
		\end{tabular}
	\end{center}
	\label{table:ierr_err}
	\vspace{-2.3em}
\end{table*}

To overcome the challenges inherent in UHD IR, we conduct an in-depth analysis of the significance of different frequency components in the restoration process through progressive frequency decoupling. 
We begin by transforming both the degraded input and ground truth (GT) into the frequency domain via discrete cosine transform (DCT), and exchange the frequency information at the (0,0) position, known as the \textit{zero frequency component} \cite{zero_frequency}. The zero-frequency component represents the direct-current information, reflecting the global and average characteristics of the image.
Reconstruction via inverse DCT (IDCT) reveals a  visual swap in appearance and a slightly higher PSNR for the exchanged input, highlighting the critical role of the zero-frequency component in the early stages of restoration, as illustrated in Figure \ref{fig:2a}. We then progressively expand the range of exchanged frequency components and observe that low-frequency components contribute to the recovery of structures and content, while the injection of high-frequency information refines details (Figure \ref{fig:2b}). \textit{Based on the observation, we decompose the complex UHD IR problem into three progressive stages: zero-frequency enhancement, low-frequency restoration, and high-frequency refinement, each aiming to learn the global mapping, coarse-grained
content, and fine-grained textures, respectively}.

\begin{figure*}[t]
	\centering
	\includegraphics[width=0.8\linewidth]{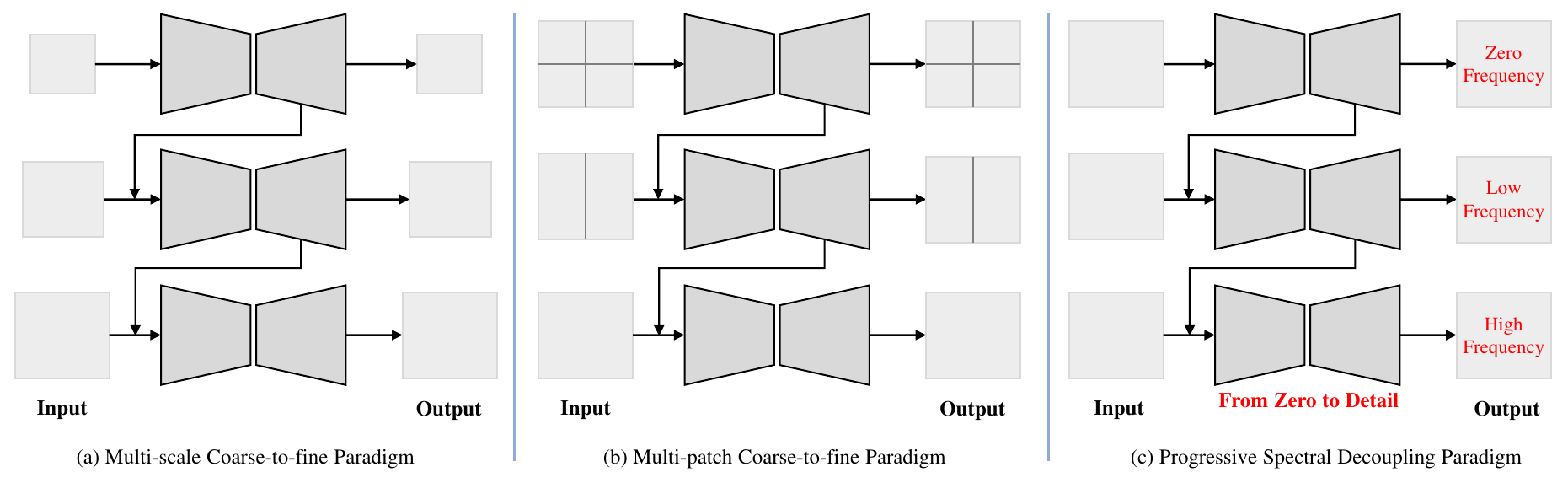}
     \vspace{-0.8em}
	\caption{Illustration of the proposed methods and the currently existing multi-stage coarse-to-fine paradigm. (a) multi-scale coarse-to-fine paradigm \cite{jiang2020multi}; (b) multi-patch coarse-to-fine paradigm \cite{zamir2021multi}; (c) our progressive spectral decoupling paradigm. Compared to previous approaches that refine images in the spatial domain without clearly defined learning objectives for each stage, our method introduces a novel paradigm that adopts a divide-and-conquer strategy in the frequency domain, performing progressive restoration from a spectral perspective.  } 
	\vspace{-0.5cm}
	\label{fig:paradigm}
\end{figure*}

Building on the insight, we propose a novel framework, termed ERR,  which addresses the challenges of UHD image restoration through progressive frequency decoupling. ERR consists of three collaborative subnetworks: the zero-frequency enhancer (ZF\textbf{E}), the low-frequency restorer (LF\textbf{R}), and the high-frequency refiner (HF\textbf{R}). Figure \ref{fig:paradigm} illustrates the comparison between our method and existing multi-stage frameworks in terms of overall paradigms. Specifically, 
ZFE enhances global image characteristics by leveraging global priors; LFR concentrates on reconstructing structural content by recovering low-frequency components; and HFR utilizes our meticulously designed frequency-windowed Kolmogorov-Arnold Network (FW-KAN) to enhance fine textures and details,  achieving high-quality image restoration. Furthermore, we present an improved model, named  IERR, which builds upon the original ERR framework with targeted architectural refinements in each subnetwork. These improvements yield a more compact, efficient, and powerful model that achieves superior restoration performance with reduced computational cost, as shown in Table \ref{table:ierr_err}.

Since the introduction of the first UHD image restoration dataset, UHDSR4K \cite{zhang2021benchmarking,li2023embedding}, a variety of UHD datasets have been developed, covering tasks such as low-light enhancement \cite{wang2023ultra}, dehazing\cite{zheng2021ultra,wang2024correlation}, deraining\cite{wang2024ultra,chen2024towards}, and deblurring\cite{deng2021multi, wang2024correlation}. These datasets are primarily constructed using two approaches: (1) selecting clean images from UHDSR4K and synthesizing paired degraded images \cite{wang2023ultra,wang2024ultra}, or (2) capturing clean images using mobile phones and generating corresponding degraded versions \cite{zheng2021ultra,deng2021multi, wang2024correlation}. However, UHDSR4K is limited in both scale and diversity, as it mainly consists of landscape images. On the other hand, datasets captured by mobile devices often suffer from limited scene variety and inconsistent image quality. As a result, the field of UHD image restoration still lacks a large-scale, high-quality dataset. To address this gap, we construct a new UHD dataset named \textbf{LSUHDIR}, comprising 82,126 high-quality UHD images carefully selected from a large-scale UHD image collection. 
Furthermore, based on LSUHDIR, we establish two benchmarks targeting UHD image denoising and JPEG artifact removal, providing a new foundation for advancing research in UHD image restoration.

In summary, the main contributions of this paper are summarized as follows:

\begin{itemize}
	\item We conduct an in-depth analysis of UHD restoration from a progressive spectral perspective, decomposing the complex UHD restoration problem into three progressive stages: zero-frequency enhancement, low-frequency restoration, and high-frequency refinement.
	
	\item Building on this insight, we propose a novel framework, termed \textbf{ERR}, which consists of three sub-networks: the zero-frequency enhancer (ZF\textbf{E}), the low-frequency restorer (LF\textbf{R}), and the high-frequency refiner (HF\textbf{R}).  Furthermore, we present an improved variant, IERR, a more compact, efficient, and powerful model that achieves superior restoration performance with lower computational cost. 
	
	\item For the ZFE, we design a global perception transformer block (GPTB) to more effectively capture global representations. For the HFR, we develop a frequency-windowed KAN (FW-KAN) to refine fine-grained information, thereby enhancing image details and textures.

    \item We construct a large-scale UHD image dataset, LSUHDIR, which contains 82,000 high-quality UHD images selected through a rigorous filtering process. Based on LSUHDIR, we further establish two benchmarks targeting UHD image denoising and JPEG compression artifact removal.

\end{itemize}

This paper expands upon our conference paper published in CVPR 2025~\cite{zhao2025zero}. 
Extensions are presented in various aspects. \textbf{In Method: (1)} We design a global prior projector (GPP) for ZFE to enhance global prior information. Additionally, we expand the downsampling space of ZFE from 8× to 16× and empirically validate its effectiveness. \textbf{(2)} We introduce a local enhancement module (LEM) for LFR to strengthen its capability in learning localized information. Meanwhile, we reduce the number of residue state space blocks (RSSBs) to maintain efficiency. \textbf{(3)} We design a zigzag reordering (ZR) strategy  for HFR to enhance its capability in frequency-domain learning. \textbf{In Benchmark: (4)} We construct a large-scale UHD image dataset, LSUHDIR, which contains 82,000 high-quality UHD
images. \textbf{(5)} Based on our LSUHDIR, we establish two benchmarks targeting UHD image denoising  and JPEG compression artifact removal. \textbf{In Experiment: (6)} Extensive experiments were conducted on our proposed benchmarks, UHD-Noise and UHD-JPEG. \textbf{(7)} Our method is compared with 4 new SOTA approaches on the UHD-LL, UHD-Haze, and UHD-Blur datasets. Moreover, we perform experiments on UHD demoiréing to evaluate the generalization of our method.  \textbf{(8)} We further extend our methods to general image restoration tasks and achieve SOTA    performance on three challenging tasks, including low-light enhancement, image dehazing and underwater image enhancement.   \textbf{(9)} Additional ablation studies are presented to analyze the effectiveness of our key components in IERR: the global prior projector, the local enhancement module, and the zigzag frequency windowing strategy.


\section{Related Work}\label{sec:related_works}
\begin{figure*}[t]
	\centering
	\includegraphics[width=0.95\linewidth]{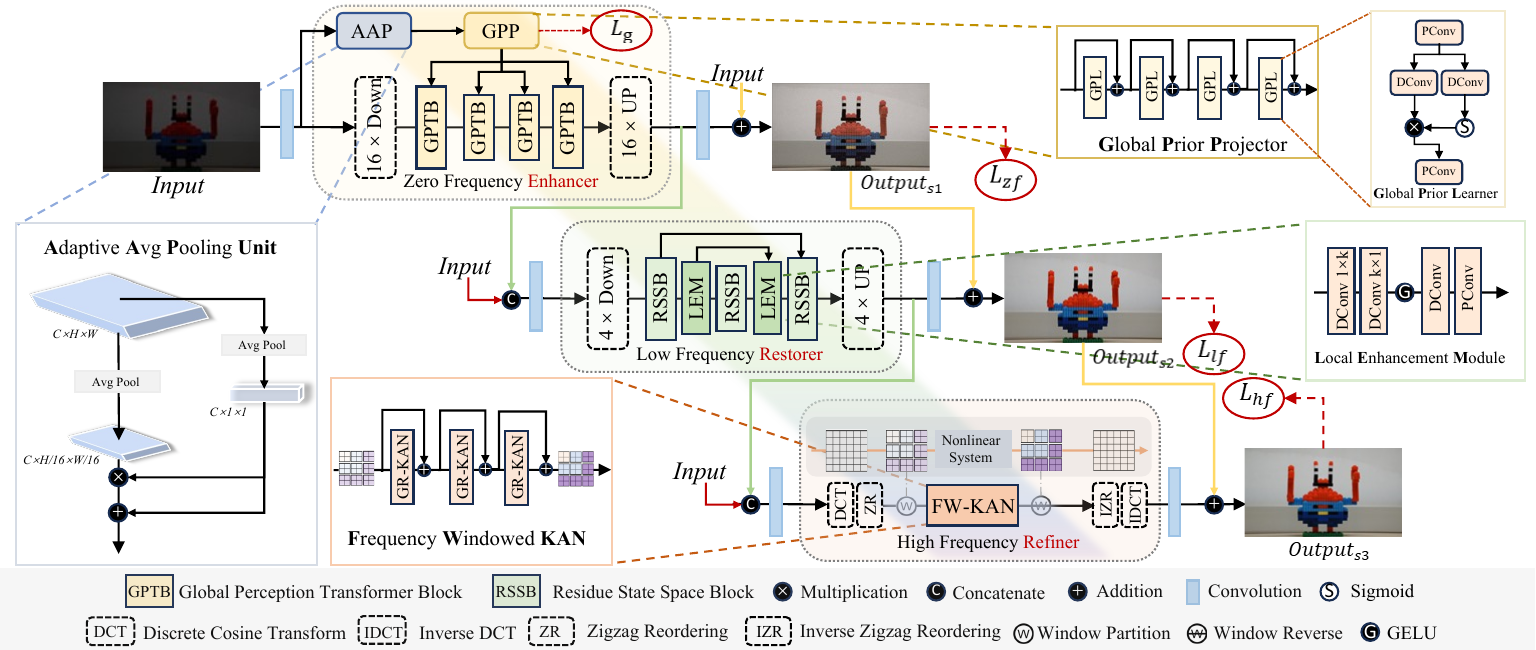}

\caption{Framework of our ERR. From a progressive spectral perspective, ERR consists of three collaborative sub-networks: the zero-frequency 
enhancer (ZFE), the low-frequency restorer (LFR), and the high-frequency refiner (HFR).  } 
	\vspace{-0.4cm}
	\label{fig:3}
\end{figure*}
\begin{figure}[t]
	\centering
	\includegraphics[width=0.95\linewidth]{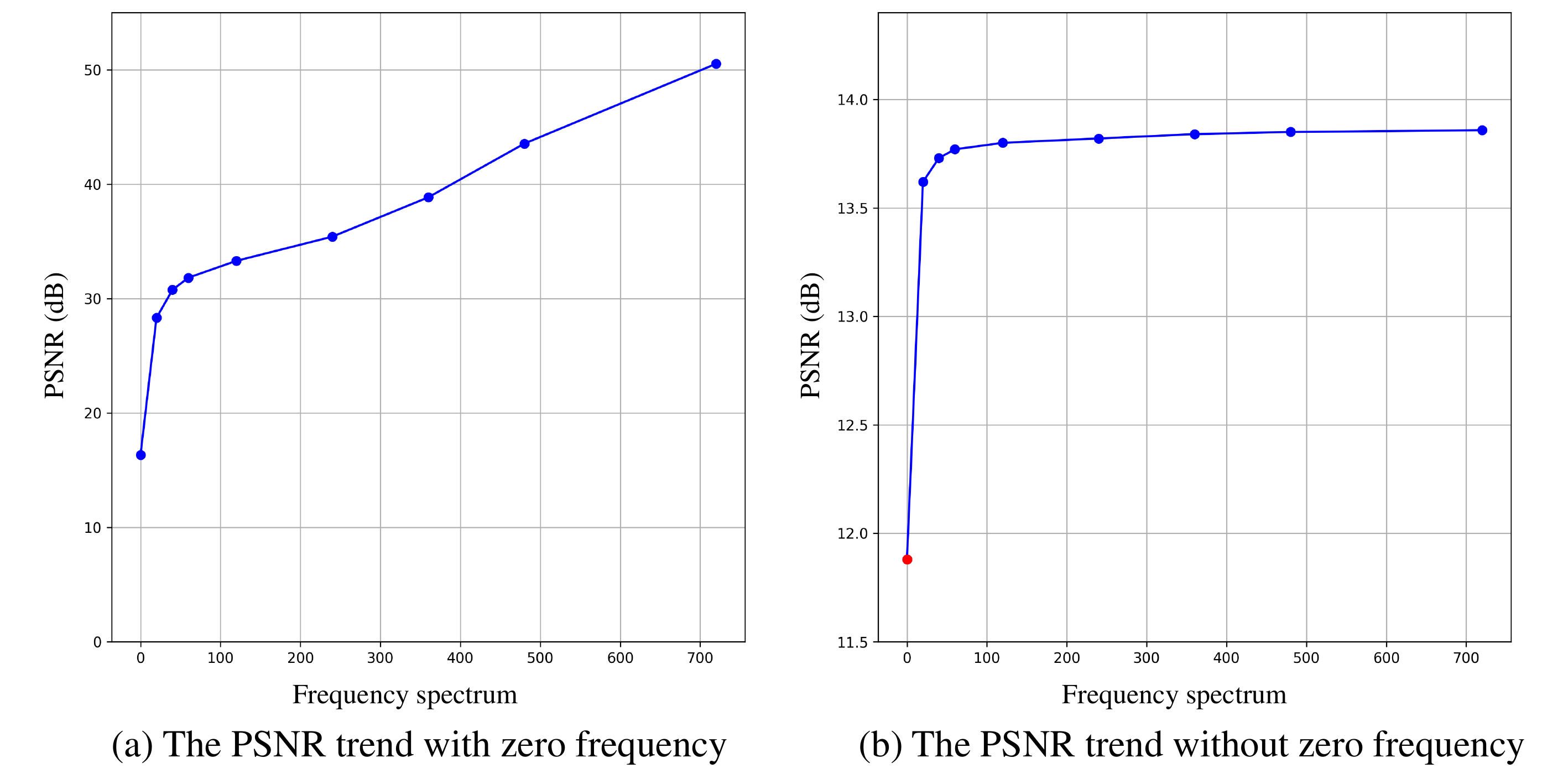}
    \caption{Analysis of the PSNR curve with spectrum exchange strategy. The growth trend (a) when initiating spectrum exchange from the zero-frequency component, (b) without exchanging the zero-frequency component.  } 
	\vspace{-0.4cm}
	\label{fig:zero_trend}
\end{figure}
\begin{figure*}[t]
	\centering
	\includegraphics[width=0.9 \linewidth]{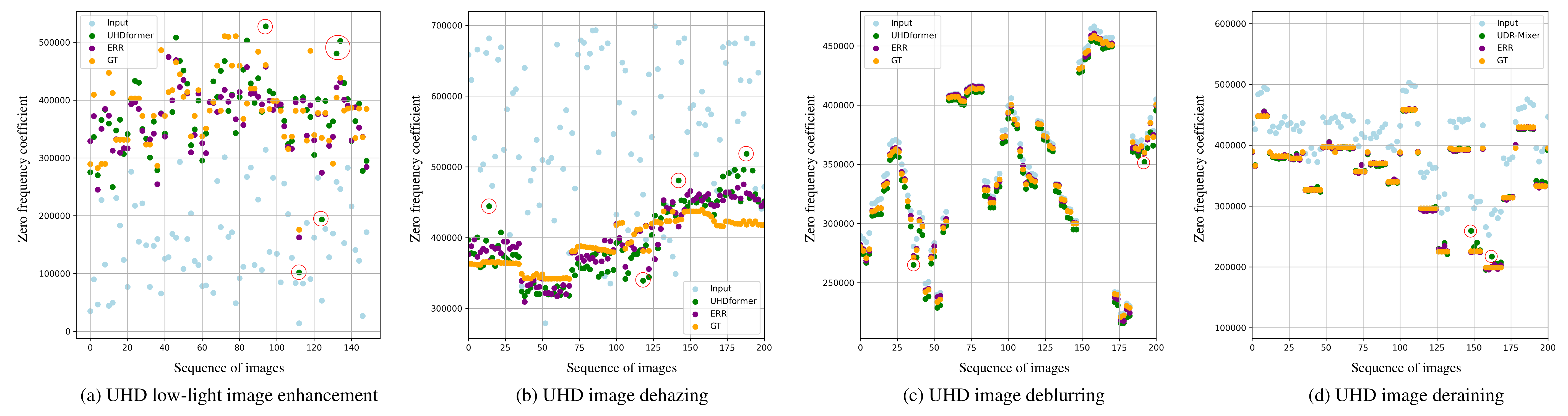}
	
    \caption{Analysis of Zero-Frequency Coefficients across Various Tasks and Methods. In the figure, a larger distance from the ground truth (orange point) indicates poorer recovery of the zero-frequency coefficient. Compared with our method, previous approaches exhibit suboptimal zero-frequency reconstruction due to insufficient emphasis on zero-frequency information. Notably, the red circles highlight extreme failure cases observed in prior methods. } 
	\vspace{-0.4cm}
	\label{fig:zero_anaylsis}
\end{figure*}

\subsection{Image Restoration}

Deep learning-based approaches have demonstrated remarkable success across various computer vision domains, particularly in image restoration (IR) tasks \cite{liang2021swinir,zamir2022restormer,zhou2024adapt,zhou2026refineanything}. These methods are generally divided into CNN-based and Transformer-based categories. Early CNN models such as SRCNN \cite{dong2015image} for super-resolution and DnCNN \cite{zhang2017beyond} for denoising pioneered the development of CNN-based IR solutions, significantly enhancing the representational strength of deep networks. For the Transformer, SwinIR \cite{liang2021swinir} and Restormer \cite{zamir2022restormer} were proposed to address diverse restoration tasks, giving rise to a wave of Transformer-based designs. Current state-of-the-art (SOTA) strategies mainly focus on improving performance through increasingly intricate network architectures \cite{li2023efficient,zhang2024distilling,lu2024robust,gao2025eraseanything,li2025set,zhou2025dragflow,DBLP:conf/cvpr/0002FZL00Z24,wang2026exposing,wei2023efficient,wang2025fast,wang2025runawayevil}. Nonetheless, the extremely high resolution and dense pixel information in UHD images pose significant challenges, often limiting the effectiveness of these  models in UHD settings, and thus hindering deployment of UHD imaging technologies.

\subsection{UHD Image Restoration }

With the increasing adoption of UHD imaging, the area of UHD restoration is gaining considerable traction. Zheng et al. \cite{zheng2021ultra} proposed a multi-guided bilateral upsampling framework for UHD dehazing. UHDFour \cite{li2023embedding} reduced the resolution of UHD images by a factor of 8, thereby enabling full-scale processing on edge platforms. UHDformer \cite{wang2024correlation} utilizes high-resolution cues to support low-resolution restoration. UDR-Mixer \cite{chen2024towards} enhanced spatial features in low-resolution images via frequency-domain modulation. These techniques \cite{zheng2021ultra,li2023embedding,wang2024correlation,chen2024towards} generally follow a unified approach: learning from scaled-down data to alleviate computational demands. However, this \textit{downsampling-enhancement-upsampling paradigm} inevitably results in the loss of essential information \cite{yu2024empowering}. Moreover, the complexity of UHD images—with \textit{ultra-high resolution, rich content, and complex structures}—poses major challenges for restoration, making it difficult for existing methods to achieve both efficiency and high quality \cite{yu2024empowering}.

\subsection{Frequency Learning}

A growing body of research \cite{qin2021fcanet,fritsche2019frequency,mao2023intriguing,zou2024wave,chen2023diffusion,chen2025ragd,du2025textcrafter,chen2025dip,chen2024adaptive,lu2025does,zhou2024general,zhao2024toward,cui2023image} has incorporated frequency-domain information into various vision tasks. DSGAN \cite{fritsche2019frequency} disentangled low- and high-frequency components of images through low-pass and high-pass filters. LaMa \cite{suvorov2022resolution} adopted frequency convolution for the task of image inpainting. DeepRFT \cite{mao2023intriguing} employed a lightweight res-fft-relu-block to address image deblurring. Fourmer \cite{zhou2023fourmer} utilized Fourier transform to capture long-range dependencies in image restoration. To better harness the diverse characteristics of frequency components, several recent studies \cite{zhao2024wavelet,jiang2023low,jiang2023dawn,zou2024wave} adopted discrete wavelet transform (DWT) for signal separation in IR tasks. Unlike previous frequency approaches, we decomposed the complex UHD restoration problem into three stages from a progressive frequency perspective,  focusing on learning the global mapping, coarse-grained 
content, and fine-grained textures, respectively.


\section{Methodology}

\begin{figure}[t]
	\centering
	\includegraphics[width=0.9 \linewidth]{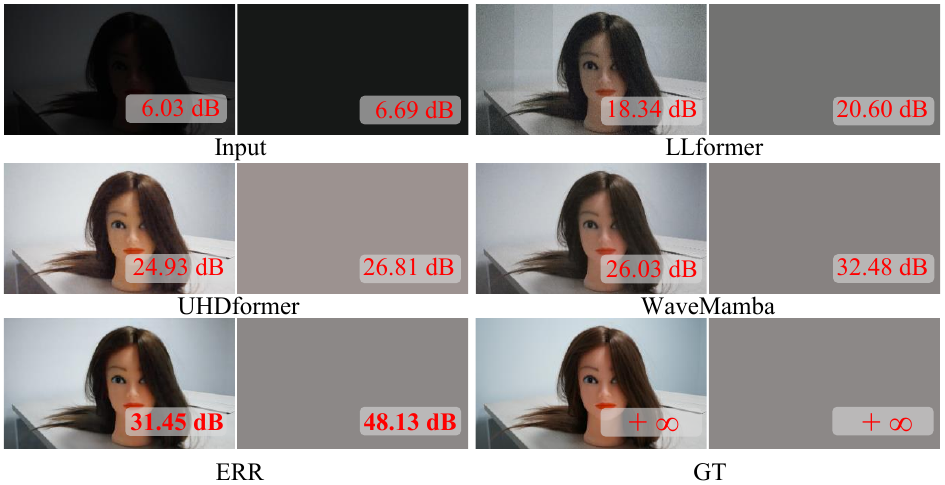}
     \vspace{-0.8em}
	\caption{Visual 
comparison of zero-frequency components. Our ERR achieves a significant global enhancement.} 
	\vspace{-0.5cm}
	\label{fig:zero_visual}
\end{figure}

\subsection{Overall Framework}

Given a degraded UHD image, the objective is to train a network that can produce a high-quality UHD output by effectively removing degradation artifacts. The overall architecture of ERR, illustrated in Figure~\ref{fig:3}, comprises three collaborative subnetworks: the zero-frequency enhancer (ZFE), the low-frequency restorer (LFR), and the high-frequency refiner (HFR). Specifically, ZFE focuses on capturing global representations within a low-resolution domain. To reconstruct the essential image content, LFR operates in a mid-resolution space to recover low-frequency components. Finally, HFR utilizes our proposed frequency-windowed Kolmogorov-Arnold Network (FW-KAN) at full resolution to refine details and textures, enabling the restoration of visually rich and high-fidelity images.

\subsection{Motivation and Analysis} 
\noindent \textbf{Discrete Cosine Transform}. Frequency decoupling and analysis have been widely applied across various fields. We primarily use the discrete cosine transform (DCT) to analyze and decompose UHD images. Given an input $x \in \mathbb{R}^{H\times W\times C}$, whose spatial shape is $H\times W$, the DCT transform $\mathcal{D}$ that converts the input $I$  to the
frequency space $F$ can be expressed as:

{
\footnotesize
\begin{align}
   \mathcal{D}(x)(u,v) &= F(u,v) \notag \\
   &= \sum_{h=0}^{H-1} \sum_{w=0}^{W-1} x(h,w) \cdot \cos\frac{(2h+1)u\pi}{2W} \cdot \cos\frac{(2w+1)v\pi}{2H},
\end{align}
}where $h, w$ and $u, v$ stand for the coordinates in the RGB color space and in the frequency space. $\mathcal{D}^{-1}$ denotes the inverse discrete cosine transform (IDCT).

\noindent \textbf{The observations}. To cope with the complexities of UHD images, we analyze and explore from progressive frequency decoupling perspective. Figure \ref{fig:2} illustrates our observations schematically. Firstly, we investigate the importance of the zero-frequency component, which can be represented as $\mathcal{D}(x)(0,0)$. The zero-frequency component refers to the direct-current information \cite{zero_frequency}, reflecting the global and average characteristics of the image. We exchange the zero-frequency components between the degraded image (input) and the GT, obtaining the exchanged input and GT. The exchanged input retains a perfect zero-frequency component while containing degraded non-zero frequency elements, while the exchanged GT is its inverse. As shown in the two cases in Figure \ref{fig:2a}, the quality of the exchanged input is significantly higher than that of the exchanged GT. Therefore, even if perfect non-zero frequency components can be learned (e.g., the exchanged GT), satisfactory results cannot be achieved. This observation suggests that \textit{in the early stages of restoration, a stronger focus should be placed on learning the zero-frequency component, i.e., capturing the global mapping}. We sequentially expand the range of frequency components, exchanging the first k levels, which can be expressed as $\mathcal{D}(x)(:k,:k)$. We observe two trends in the changes of image quality. As the low-frequency information is progressively filled, the content and structural information of the coarse-grained level are gradually restored, while the injection of high-frequency information refines the details and textures of the fine-grained level. The PSNR value and visual quality of the input gradually improve, as illustrated in Figure \ref{fig:2b}. This step-by-step  process from zero to detail 
implies a gradual enhancement in image quality; in contrast, the reverse process fails to achieve high-quality results.

\noindent \textbf{The core motivation}. The above observations highlight two key findings: 1) The evolution of image appearance reflects a three-stage restoration process—global enhancement, low-frequency filling, and high-frequency injection. 2) The increasing 
PSNR trend indicates a progressive restoration process, illustrating the ideal progression of the model’s learning, as 
shown in Figure \ref{fig:zero_trend} (a). Figure \ref{fig:zero_trend} (b) presents the trend without zero-frequency transformation, exhibiting a suboptimal  growth pattern, which demonstrates the critical role of zero-frequency. Based on this observation and analysis, our goal is to decompose the complex UHD restoration problem into three progressive stages: \textit{zero-frequency enhancement, low-frequency restoration, and high-frequency refinement, each focusing on learning the global mapping, coarse-grained content, and fine-grained textures, respectively.} 

\noindent \textbf{The analysis for zero frequency coefficient}. As illustrated in Figure \ref{fig:zero_anaylsis}, we conduct a visual analysis of zero-frequency components across multiple UHD image restoration tasks. Compared with recent state-of-the-art methods, our ERR achieves more consistent and stable recovery of the zero-frequency components. Prior methods, which often overlook the importance of zero-frequency information, exhibit inferior performance in certain cases, as highlighted by the red circles in Figure \ref{fig:zero_anaylsis}. Furthermore, we provide an additional comparison in Figure \ref{fig:zero_visual}, where the visualization of zero-frequency components further demonstrates the superior global enhancement capability of our method.

\vspace{-0.8em}
\begin{figure}[t]
	\centering
	\includegraphics[width=0.9 \linewidth]{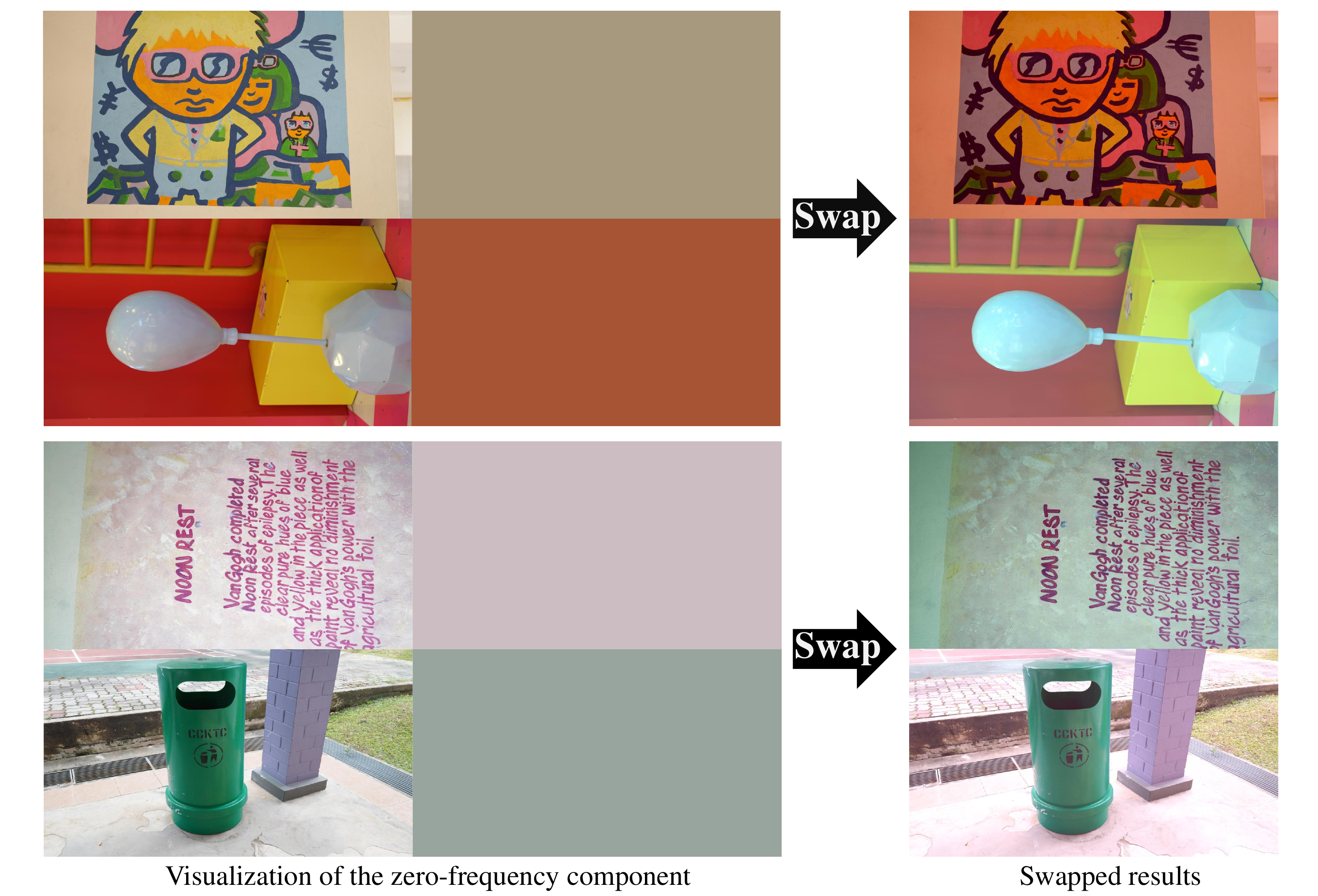}
     \vspace{-0.8em}
	\caption{Zero-frequency visualization. We swap the zero-frequency components between two images with entirely different semantic content and observe that the resulting changes occur primarily in \textit{global color and luminance}, while the semantic structures remain unaffected.} 
	\vspace{-0.2cm}
	\label{fig:zero_show}
\end{figure}

\subsection{Zero Frequency Enhancer }
 
ZFE aims to learn the global mapping in a low-resolution space and is composed of three key components: the Adaptive Average Pooling (AAP) unit, the Global Prior Projector (GPP), and the Global Perception Transformer Block (GPTB). The AAP unit is designed to extract global priors from the full-resolution space. The GPP further enhances the  captured global priors, making them more informative and representative. By leveraging the extracted prior, the GPTB can more effectively facilitate global representation learning. IERR introduces two main improvements at this stage: 1) the incorporation of the GPP along with a global prior loss to supervise global information learning; 2) the use of a larger downsampling scale to improve computational efficiency.

\noindent \textbf{Why can ZFE operate effectively in an extremely low-resolution space?} As shown in Figure \ref{fig:zero_show}, visualizations of zero-frequency components reveal that they primarily encode global properties such as luminance and color. To further verify this, we exchange the zero-frequency components between two semantically unrelated images. The exchanged results exhibit noticeable changes in global color tones while preserving semantic and structural content. This indicates that zero-frequency information governs overall appearance without affecting high-level semantics. Based on this insight, IERR increases the downsampling ratio from $\times$8 to $\times$16, enabling more efficient processing while retaining the ability to learn global representations.

\begin{figure}[t]
	\centering
	\includegraphics[width=0.9 \linewidth]{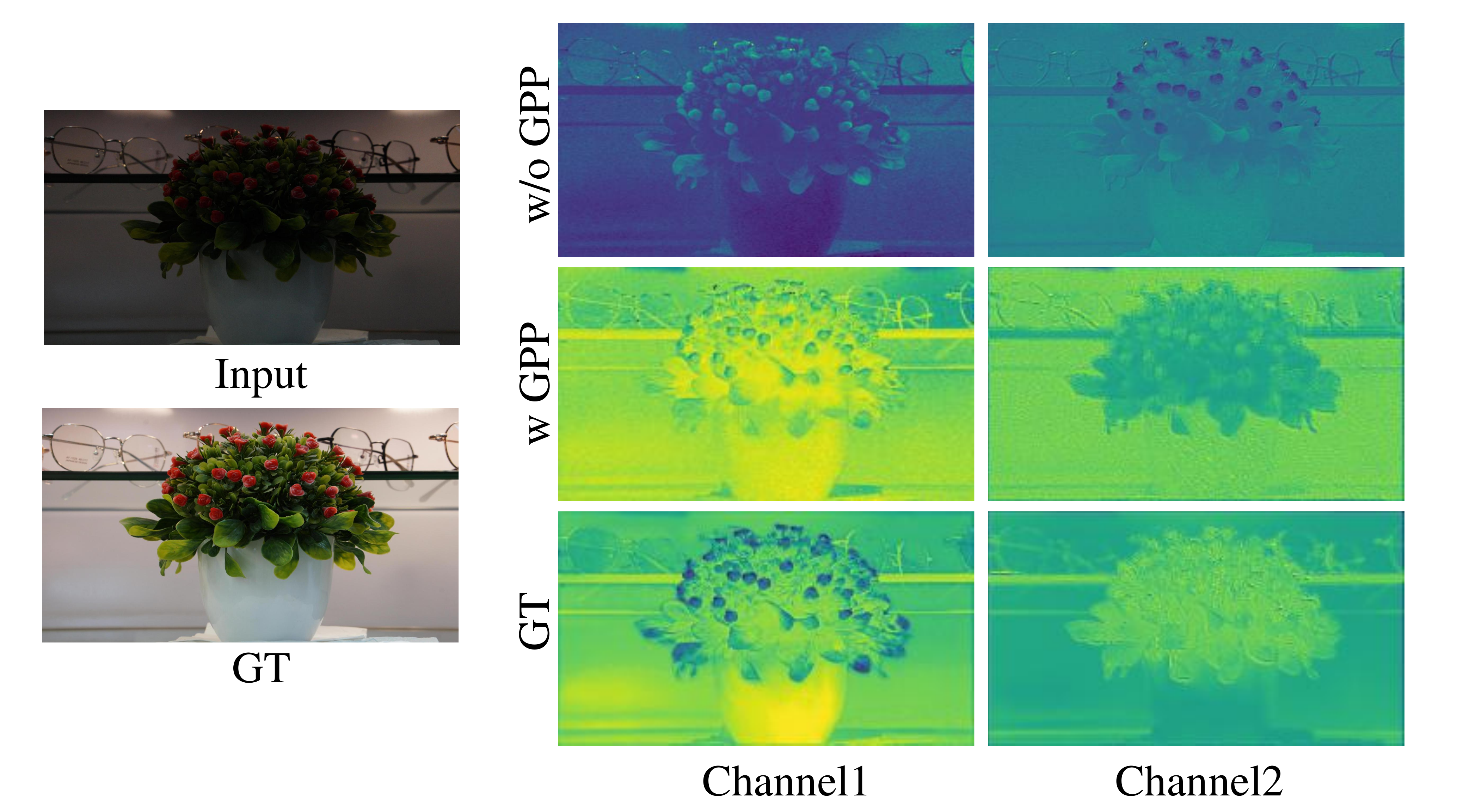}
     \vspace{-0.8em}
	\caption{Feature visualizations for the ablation study of GPP. GT denotes that the ground truth is used as input, and its corresponding feature maps are visualized for reference. } 
	\vspace{-0.5cm}
	\label{fig:feature_map}
\end{figure}

\begin{table}[t]
\centering
\caption{Quantitative analysis of the zero-frequency component on the UHD-LL for the ablation study of GPP.}
\label{tab:zero_frequency_Quanti}
\begin{tabular}{lc}
\toprule
Method & Zero-Frequency PSNR $\uparrow$  \\
\midrule
w/o GPP  & 28.15 \\
w/ GPP & 29.18 \\
\bottomrule
\end{tabular}
\vspace{-0.3cm}
\end{table}

\noindent \textbf{AAP unit}. The zero-frequency component represents the direct current (DC) information, which reflects the global and average characteristics of the UHD image. Consequently, we employ average pooling (AvP) in the full-resolution space to capture global prior. Given an input feature map $x\in\mathbb{R}^{H\times W\times C}$, we aim to extract global features while preserving local characteristics. To achieve the target, we employ a combination of global AvP and local AvP. The AAP unit can be expressed mathematically as:
\begin{align}
   \mathcal{AAP}(x) = AvP_{1,1}(x) \odot AvP_{\frac{H}{16}, \frac{W}{16}}(x) + AvP_{1,1}(x),
\end{align}
where $AvP_{1,1}$ represents the global AvP with a pooling size of \( (1, 1) \), and 
$AvP_{\frac{H}{16}, \frac{W}{16}}(x)$ denotes the local AvP with a pooling size of \( \left( \frac{H}{16}, \frac{W}{16} \right) \).  $\odot$  indicates the element-wise  multiplication. 

\noindent \textbf{GPP}. Since the extracted global prior is directly derived from degraded input features, it inevitably contains significant corruption. Directly fusing such priors into the GPTB impairs the learning of global representations. To address this, we introduce a global prior projector (GPP) between the AAP and GPTB to enhance the global prior and better guide global mapping. As shown in Figure \ref{fig:feature_map}, without GPP, the fused feature maps tend to be dark and lack structure. With GPP, the features become more similar to those obtained using ground-truth as input, indicating that GPP effectively enhances the global prior. Table \ref{tab:zero_frequency_Quanti} further quantitatively demonstrates that the model equipped with the GPP achieves a significantly higher zero-frequency PSNR compared to the variant without it. The GPP consists of stacked gated convolution blocks, and we design a global prior loss to explicitly supervise the learning of GPP:
\begin{equation}
G=GPP(\mathcal{AAP}(x)),
\end{equation}
\begin{equation}
    \mathcal{L}_{g}=||Conv(G)-\mathcal{AAP}(GT)||_1,
\end{equation}
where $G \in\mathbb{R}^{\frac{H}{16}\times\frac{W}{16}\times C}$ refers to the global prior, and $Conv$ is employed to adjust the feature channels.

\begin{figure}[t]
	\centering
	\includegraphics[width=0.98\linewidth]{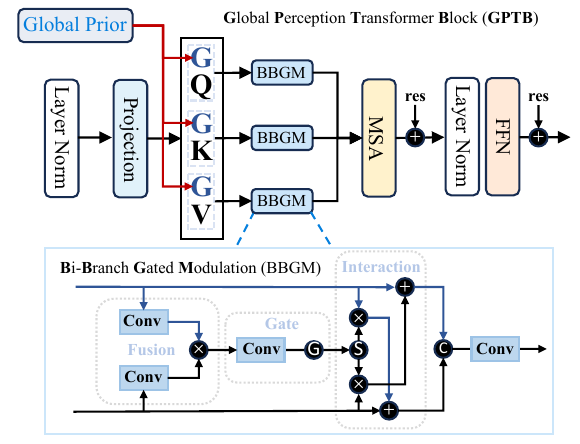}
     \vspace{-0.8em}
	\caption{Architecture of the global perception transformer block (GPTB).  } 
	\vspace{-0.5cm}
	\label{fig:4}
\end{figure}
\noindent \textbf{GPTB}. The low-resolution features $x_{\mathrm{low}}\in\mathbb{R}^{\frac{H}{16}\times\frac{W}{16}\times C}$  are obtained from an input $x$ via 16x downsampling. Then, $x_{\mathrm{low}}\in\mathbb{R}^{\frac{H}{16}\times\frac{W}{16}\times C}$ is sent to several GPTBs to learn the global mapping. Figure \ref{fig:4} shows the detail architecture of GPTB, where bi-branch gated modulation (BBGM) is designed to integrate low-resolution features with global priors.  The computation in the GPTB can be denoted:
\begin{align}
   Q,K,V= \mathcal{S}(Projection(LN(x_{\mathrm{low}}^{i-1}) )),
\end{align}
\begin{align}
   \hat{Q}, \hat{K}, \hat{V} = \mathcal{B}(G, Q), \ \mathcal{B}(G, K), \ \mathcal{B}(G, V),
\end{align}
\begin{equation}
    \hat{x}=MSA(\hat{Q}, \hat{K}, \hat{V})+x_{\mathrm{low}}^{i-1},
\end{equation}
\begin{equation}
   x_{\mathrm{low}}^{i}=FFN( LN( \hat{x}))+\hat{x},
\end{equation}
where  LN, MSA, and $\mathcal{S}$ refer to layer normalization, multi-head self-attention, and split operation, respectively.  $x_{\mathrm{low}}^{i-1}$ represents the input embeddings of the current GPTB. The BBGM $\mathcal{B}$ comprises three part—fusion, gating, and interaction—that collectively ensure a comprehensive integration of features. The BBGM can be expressed as follows:
\begin{equation}
    W_{a}, W_{b} = \mathcal{S} \left( \delta \left( \mathrm{Conv} \left( \mathrm{Conv}(G) \odot \mathrm{Conv}(F) \right) \right) \right),
\end{equation}
\begin{equation}
    \hat{F} = \mathcal{C} \left[ F + W_{a} \odot G, G + W_{a} \odot F \right],
\end{equation}
where $F$ and $\hat{F}$ represent the original embedding and the features fused with the global prior $G$. $\mathcal{C}$ and $\delta$ refer to the concat operation and Gelu function. 

\noindent \textbf{Regularization}. In the first stage, we primarily constrain the zero-frequency component of the generated image, focusing on learning global representation. Given that the output of the first stage is $O_{s1}$, zero-frequency regularization $\mathcal{L}_{zf}$ can be mathematically expressed as:
\begin{equation}
    \mathcal{L}_{zf}=||\mathcal{D}(O_{s1})(0,0)-\mathcal{D}(GT)(0,0)||_1.
\end{equation}
 \vspace{-0.3cm}

\subsection{Low Frequency Restorer }
\begin{figure}[t]
	\centering
	\includegraphics[width=0.98\linewidth]{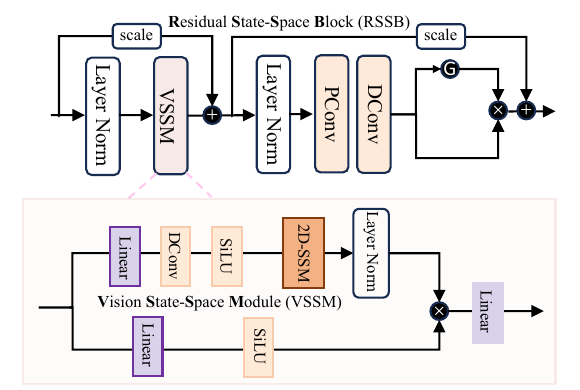}
     \vspace{-0.8em}
	\caption{Architecture of the residual state space block (RSSB). } 
	\vspace{-0.2cm}
	\label{fig:5}
\end{figure}

The LFR learns low-frequency representations within a mid-resolution space to efficiently restore content information. The mid-resolution features $x_{\mathrm{mid}}\in\mathbb{R}^{\frac{H}{4}\times\frac{W}{4}\times C}$ are obtained via 4× downsampling. The core component of LFR is the residue state space block (RSSB), also referred to as the Mamba block \cite{guo2025mambair}, which enables efficient modeling of long-range dependencies with low computational overhead. At this stage, IERR integrates a local enhancement module (LEM) designed to improve the modeling of local features. This module mainly leverages convolution operations, and the use of large-kernel convolution helps to broaden the receptive field. As illustrated in Figure~\ref{fig:5}, the RSSB can be mathematically expressed as:
\begin{equation}
    x^{\prime}=\text{VSSM}(\text{LN}(x_{\mathrm{mid}}^{i-1}))+s\cdot x_{\mathrm{mid}}^{i-1},
\end{equation}
\begin{equation}
    x_{\mathrm{mid}}^{i}=\text{PC}(\delta_{g}( \text{DC}(\text{PC}(\text{LN}(x^{\prime}))))+s'\cdot x^{\prime}),
\end{equation}
where $PC$ and $DC$ represent point-wise and depth-wise convolution, respectively. $\delta_{g}$ is the function of non-linear gate, similar
to SimpleGate,  dividing the input
along the channel dimension into two features $\mathbf{F}_{1},\mathbf{F}_{2}\in\mathbb{R}^{H\times W\times\frac{C}{2}}$. The output is then calculated by $\delta_{g}(F)=\delta(\mathbf{F}_{1}) \cdot \mathbf{F}_{2}$. The detailed description of the vision state-space module (VSSM) \cite{guo2025mambair} can be found in the supplementary materials.

\noindent \textbf{Regularization}. In this stage, our objective is to learn the low-frequency components,  with a primary focus on the restoration of coarse-grained structures and content. Given that the output of this stage is $O_{s2}$ and frequency cutoff $k$ is hyper parameter, low-frequency regularization $\mathcal{L}_{lf}$ can be mathematically expressed as:
\vspace{-0.3cm}

\begin{equation}
\begin{split}
    \mathcal{L}_{lf} = &\sum_{i=0}^{k} \sum_{j=0}^{k} \left\| \mathcal{D}(O_{s2})(i, j) - \mathcal{D}(GT)(i, j) \right\|_1, \hfill \\
    &\hfill \hspace{8.25 em} \text{where } (i,j) \neq (0,0).
\end{split}
\end{equation}

\vspace{-0.6em}
\subsection{High Frequency  Refiner }
In the high-frequency refinement stage, HFR employs the proposed FW-KAN to refine image details and textures in the full-resolution space, thereby enabling high-quality reconstruction. 

\begin{figure}[t]
	\centering
	\includegraphics[width=1 \linewidth]{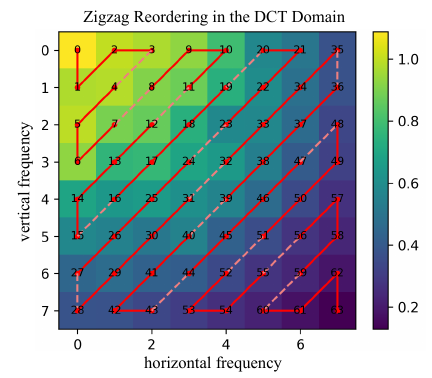}
     \vspace{-2.5 em}
	\caption{Zigzag reordering for IERR. The red line denotes the zigzag scanning path from low to high frequencies, where each solid red segment indicates a grouped window.  } 
	\vspace{-0.1cm}
	\label{fig:zigzag}
\end{figure}

\noindent \textbf{Zigzag reordering for IERR}. To further enhance frequency-domain learning, we introduce a zigzag reordering (ZR) strategy tailored for HFR. The previous conference version directly partitioned the DCT feature space into windows without accounting for the underlying frequency distribution, potentially resulting in suboptimal grouping of frequency components. Inspired by the zigzag scanning mechanism used in JPEG compression, which arranges low-frequency coefficients at the beginning and high-frequency ones at the end, we adopt a similar strategy to reorder frequency components in the DCT domain. As illustrated in Figure \ref{fig:zigzag}, the reordering results in a coherent low-to-high frequency arrangement, which subsequently guides the design of frequency-aware window partitioning. 

\begin{figure}[t]
	\centering
	\includegraphics[width=0.9 \linewidth]{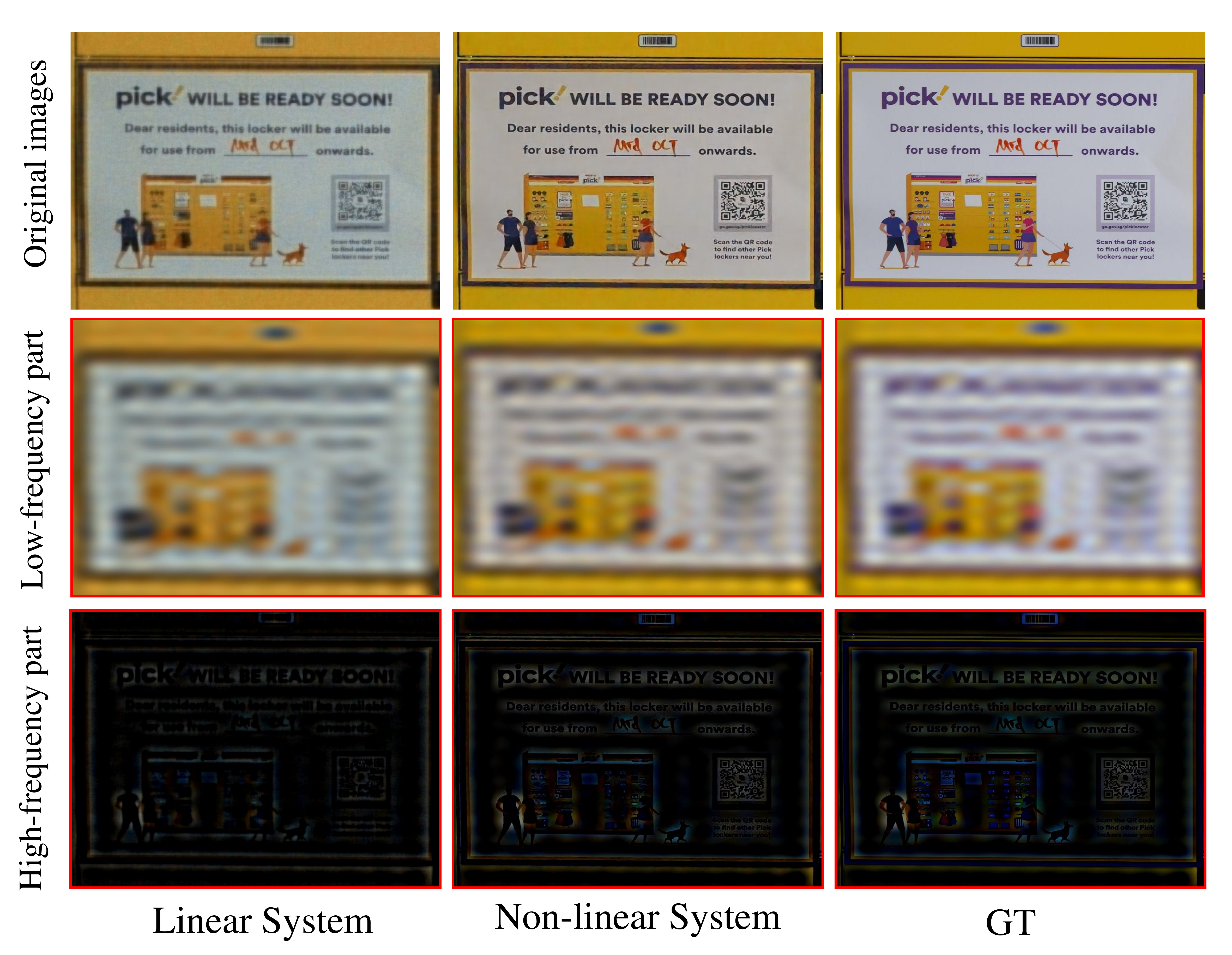}
     \vspace{-0.8em}
	\caption{Visual comparison with linear system. 
    The visual results suggest that the non-linear functions mainly inject high-frequency information. } 
	\vspace{-0.5cm}
	\label{fig:nonliear}
\end{figure}

\noindent \textbf{The motivation for KAN}. Recent work on explainability in deep learning \cite{deng2024exploring} introduces an insightful perspective: the linear system primarily acts as a low-frequency learner, whereas \textit{the non-linear system injects high-frequency information}. Inspired by this observation, we further conduct an empirical analysis on UHDFormer by removing all non-linear activation functions, including GELU, softmax and other non-linear functions. As shown in Table~\ref{tab:nonlinear} and Figure~\ref{fig:nonliear}, the results validate that the non-linear function primarily injects high-frequency information.  Recently, KAN \cite{liu2024kan} has become well-known as a learnable non-linear operator with powerful non-linear expression capabilities, which motivates us to explore it as the core operator in our final stage.

\begin{table}[t]
	
    \caption{UHDFormer is adopted as the baseline for the non-linear system, and Linear system means UHDformer without all non-linear functions. The difference on UHD-LL  \cite{li2023embedding} suggests that the non-linear functions mainly injects high-frequency information.}
	\centering
	
	\resizebox{0.95\linewidth}{!}{
		\begin{tabular}{c| c c c  c |c c}
			\specialrule{1.2pt}{0.2pt}{1pt}
			Method&\multicolumn{2}{c}{Linear system} &\multicolumn{2}{c}{Non-linear System} &  \multicolumn{2}{|c}{Difference~} \\
            \cmidrule(lr){1-1}
			\cmidrule(lr){2-5}
			\cmidrule(lr){6-7}
			Frequency& Low& High & Low & High & Low  & High \\
			\midrule
			PSNR $\uparrow$&28.70  &34.68& 27.92 & 36.82 & -0.78 & \textbf{2.14}\\
			SSIM $\uparrow$&0.9873& 0.8539 & 0.9820& 0.9081 & -0.0053 &\textbf{0.0542}\\
			
			\specialrule{1.2pt}{0.2pt}{1pt}
	\end{tabular}}
    \vspace{-1.8em}
	\label{tab:nonlinear}
	
\end{table}

\noindent \textbf{FW-KAN}.  
While group-rational KAN (GR-KAN) \cite{yang2024kolmogorov} mitigates the high complexity and optimization difficulties of the original KAN \cite{liu2024kan} through the introduction of rational activation functions and variance-preserving initialization, its application in the UHD original resolution space still demands significant memory resources. Furthermore, the pixel density of UHD images renders high-frequency details extraordinarily complex. To alleviate these issues, we adopt a window partition (WP) in the DCT spectrum to partition high- and low-frequency information into blocks, which not only reduces memory consumption but also encourages the model to concentrate more effectively on high-frequency detail learning, as illustrated in the detailed FW-KAN diagram in Figure \ref{fig:3}. It is noteworthy that in the IERR, we perform zigzag reordering (ZR) prior to WP. FW-KAN consists of multiple stacked GR-KAN. Given the original resolution feature $x_{high} \in \mathbb{R}^{H\times W\times C}$, HFR can be expressed as:
\begin{equation}
    x^{\prime}=\mathcal{D}^{-1}(\text{IZR}(\text{WR}(\text{FW-KAN}(\text{WP}(\text{ZR}(\mathcal{D})(x_{\mathrm{high}})))))),
\end{equation}
where $\mathcal{D}^{-1}$, $\text{IZR}$ and $\text{WR}$ stand for inverse DCT, inverse ZR and window reverse operation.

\noindent \textbf{Regularization}. In the final stage, our objective is to inject high-frequency information, with an emphasis on the refinement of fine-grained texture and details. Given that the output of this stage is $O_{s3}$, 
high-frequency regularization $\mathcal{L}_{lf}$ can be mathematically expressed as:
\begin{equation}
\begin{split}
    \mathcal{L}_{hf} = &\sum_{i=0}^{N} \sum_{j=0}^{N} \left\| \mathcal{D}(O_{s3})(i, j) - \mathcal{D}(GT)(i, j) \right\|_1, \hfill \\
    &\hfill \hspace{6.25 em} \text{where } (i \geq k) \lor (j \geq k),
\end{split}
\end{equation}
where $N$ denotes the maximum horizontal and vertical index of the spectrum, and $k$ represents the frequency cutoff.

\subsection{Loss }
At any stage $l$, we employ L1 and SSIM loss, and the reconstruction loss $\mathcal{L}_{rec}$ can be expressed as:
\begin{equation}
    \mathcal{L}_{rec}
=\sum_{l=1}^3\left[\mathcal{L}_{1}(O_{l},GT)+\mathcal{L}_{ssim}(O_{l},GT)\right],
\end{equation}
where $O_{l}$ represents the output at stage $l$. The total loss function $\mathcal{L}_{total}$ can be defined as: 
\begin{equation}
    \mathcal{L}_{total}
=\mathcal{L}_{rec}+\mathcal{L}_{g}+\mathcal{L}_{zf}+\mathcal{L}_{lf}+\mathcal{L}_{hf}.
\end{equation}

\section{Constructing LSUHDIR}

To address the scarcity of large-scale, high-quality datasets for ultra-high-definition (UHD) image restoration, we introduce LSUHDIR, a new benchmark comprising 82,126 high-quality UHD images. Specifically, 80,126 images are allocated for training, 1,000 for validation, and 1,000 for testing. As shown in Table~\ref{tab:compare_datasets}, LSUHDIR significantly surpasses existing popular image restoration datasets in both resolution and the number of high-quality samples.



\begin{table}[!t]
\vspace{-0.5em}
    \caption{\small{Comparison between existing datasets and our proposed LSUHDIR dataset.}}
     \vspace{-0.8em}
	\begin{center}
		\resizebox{7cm}{!}{
		\begin{tabular}{c|c|c|c}
			\hline
			\textbf{Dataset} & Year & Size & Avg.Resolution  \\
			\hline
RealSR \cite{Cai2019TowardRS} & 2019 & 595 & 1,541 $\times$ 1,302  \\
L20 \cite{Timofte2015SevenWT} & 2015 & 20 & 3,843 $\times$ 2,870  \\ 
DIV8K \cite{Gu2019DIV8KD8} & 2019 & 1,504 & 5,557 $\times$ 3,935  \\

UHDSR4K \cite{Zhang2021BenchmarkingUI}  & 2021 & 8,099 & 3,840 $\times$ 2,160  \\ 
UHDSR8K \cite{Zhang2021BenchmarkingUI} & 2021 & 2,966 & 7,680 $\times$ 4,320  \\ 
            \hline

            LSUHDIR  & 2025 & 82,126 & 5,303 $\times$ 3,535  \\
			\hline
		\end{tabular}
		}
	\end{center}
	\label{tab:compare_datasets}
    \vspace{-0.5cm}
\end{table}

\begin{figure}[t!]
    \centering
    \includegraphics[width=0.9 \linewidth]{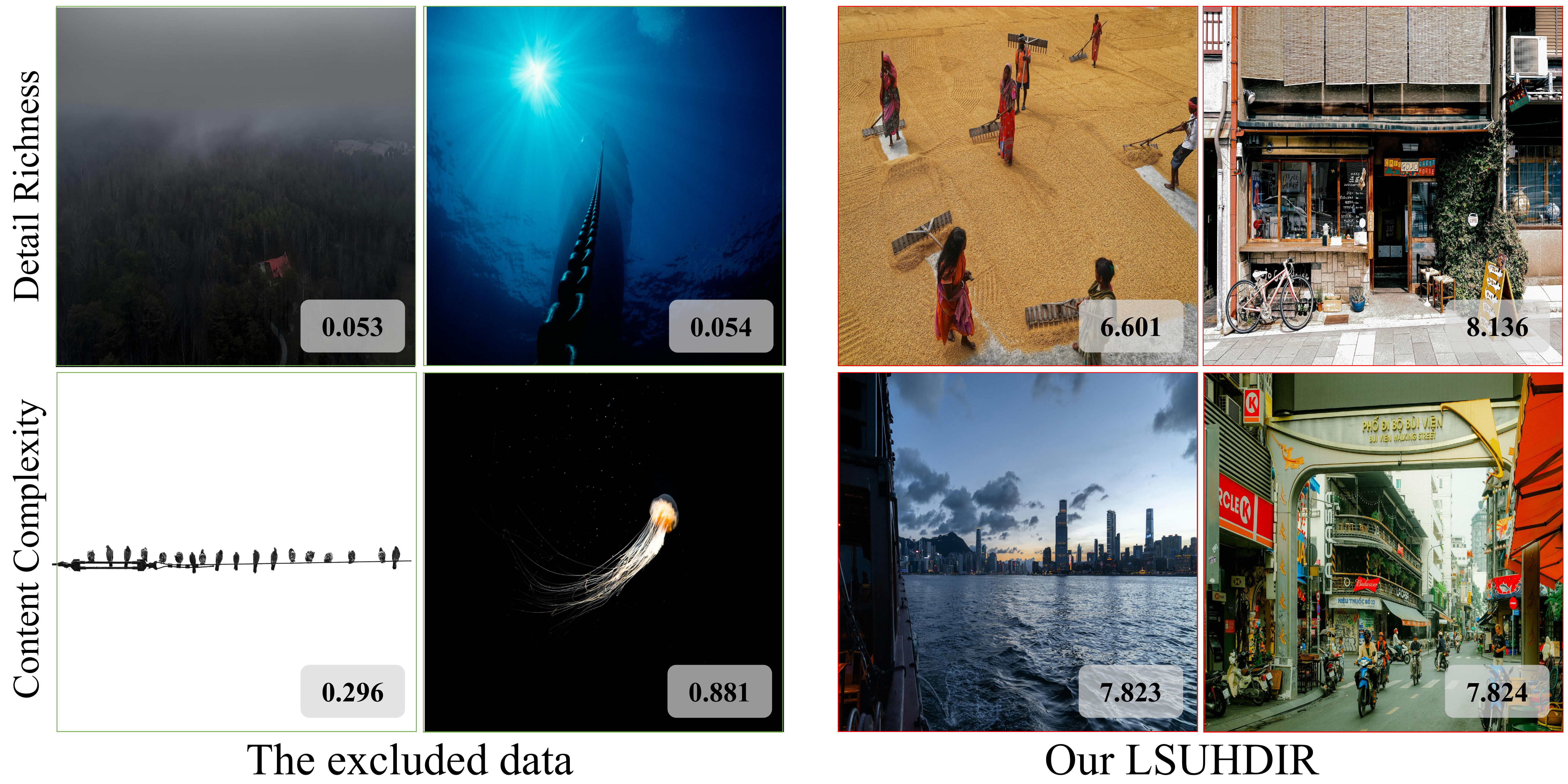}
    \vspace{-2mm}
    \caption{\textbf{Left:} We showcase representative low-quality (failure) cases with their corresponding scores in two key dimensions—detail richness and content complexity—underscoring the necessity of such filtering. \textbf{Right:} In contrast, our LSUHDIR samples demonstrate superior texture fidelity and higher semantic complexity.}
    \label{fig:LSUHDIR}
    \vspace{-4mm}
\end{figure}

\subsection{Data Collection}
Our dataset is primarily constructed from two open-license platforms, Unsplash\footnote{\url{https://unsplash.com/}} and Pexels\footnote{\url{https://www.pexels.com/}}, both of which provide images under highly permissive licenses that allow free use for both commercial and non-commercial purposes without requiring explicit permission or attribution.
These sources ensure legal compliance for dataset release and render them well-suited for building a publicly available benchmark. To guarantee content diversity, we employed more than 100 distinct search queries covering a broad spectrum of categories. These include natural landscapes, urban environments, human portraits, wildlife, and architectural structures. This strategy effectively ensures substantial variation in both semantic content and visual textures.


\begin{figure}[t!]
    \centering
    \includegraphics[width=0.9 \linewidth]{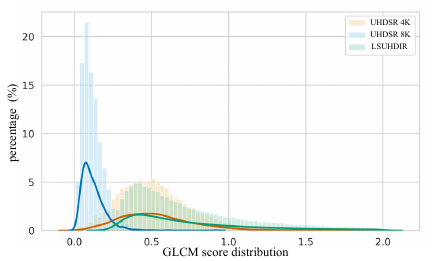}
    \vspace{-2mm}
    \caption{Comparison of the GLCM score distributions between our LSUHDIR dataset and the UHDSR4K and UHDSR8K datasets \cite{Zhang2021BenchmarkingUI}.}
    \vspace{-0.3cm}
    \label{fig:glcm}
\end{figure}
\begin{figure}[t!]
    \centering
    \includegraphics[width=0.9 \linewidth]{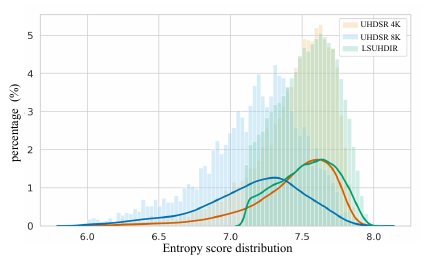}
    \vspace{-4mm}
    \caption{Comparison of the shannon entropy score distributions between our LSUHDIR dataset and the UHDSR4K and UHDSR8K datasets \cite{Zhang2021BenchmarkingUI}.}
    \vspace{-0.4cm}
    \label{fig:entrophy}
\end{figure}
\subsection{Preliminary Quality Screening.}

Unlike high-level vision tasks, constructing a high-quality image restoration dataset requires strict control over pixel-level fidelity. Web-crawled high-resolution images often suffer from various degradations, such as blur, noise, or texture sparsity. To mitigate these issues, we perform a two-stage  low-level quality screening.

\noindent\textbf{Blur and noise detection.} To remove overly blurry or excessively sharp images, we evaluate the sharpness of each image using a Laplacian-based metric. Images that are too blurry or contain excessive noise are discarded, ensuring that only images with moderate sharpness are retained.

\noindent\textbf{Flat region detection.} To identify  overly flat images, we employ the Sobel operator to measure edge density. Images with insufficient edge content are excluded from the dataset.
Through this two-stage screening, we obtain a refined candidate set $S$ with satisfactory low-level fidelity.

\subsection{High-Quality Image Assessment}
While the preliminary screening ensures basic pixel-level fidelity, achieving high-quality restoration also requires images with rich details and diverse content. Therefore, we further assess the images in $S$ based on the two characteristics, including texture richness and semantic complexity. These higher-level assessments guide the selection of images that provide more informative supervision for training restoration models, forming the next stage of the dataset. 

\noindent\textbf{Detail Richness Assessment.}  Preserving high-frequency content is critical in UHD image restoration. To evaluate the level of fine-grained details, we compute the Gray-Level Co-occurrence Matrix (GLCM) metrics, including contrast, entropy, and correlation across multiple orientations. These statistics capture spatial pixel dependencies that are indicative of texture complexity. We retain the top 50\% of images from $S$ with the highest aggregated GLCM scores, forming the detail-preserving subset $S_G$.

\noindent\textbf{Content Complexity Assessment.} Images with diverse spatial structures and semantic richness provide valuable supervision for training deep models. We measure content complexity via Shannon entropy, which reflects the diversity of pixel intensity distributions. From the same candidate set $S$, we select the top 50\% of images with the highest entropy, resulting in subset $S_E$.

\noindent\textbf{LSUHDIR Dataset.} To ensure that the final dataset consists of UHD images that are both perceptually rich and structurally diverse, we take the intersection of the two filtered subsets, $S_G \cap S_E$. This guarantees that each image meets both the high-detail and high-complexity criteria. Finally, we perform a rigorous human inspection to verify that all images comply with the high-quality standards, resulting in the final LSUHDIR dataset. Representative examples are shown in Figure~\ref{fig:LSUHDIR}. Figures~\ref{fig:glcm} and ~\ref{fig:entrophy} demonstrate that our LSUHDIR dataset exhibits significantly improved GLCM and entropy score distributions compared to the GLCM score distributions of existing datasets.

\subsection{Benchmark Construction}
Based on the curated LSUHDIR dataset, we further establish two additional benchmarks to facilitate research on specific low-level restoration tasks. The first benchmark, \textbf{UHD-Noise}, focuses on image denoising and provides high-resolution images with synthetically added noise at varying levels. The second benchmark, \textbf{UHD-JPEG}, targets the removal of JPEG compression artifacts and contains images compressed at different quality factors. 


\section{Experiments}

\begin{figure}[t]
	\centering
	\includegraphics[width=0.98\linewidth]{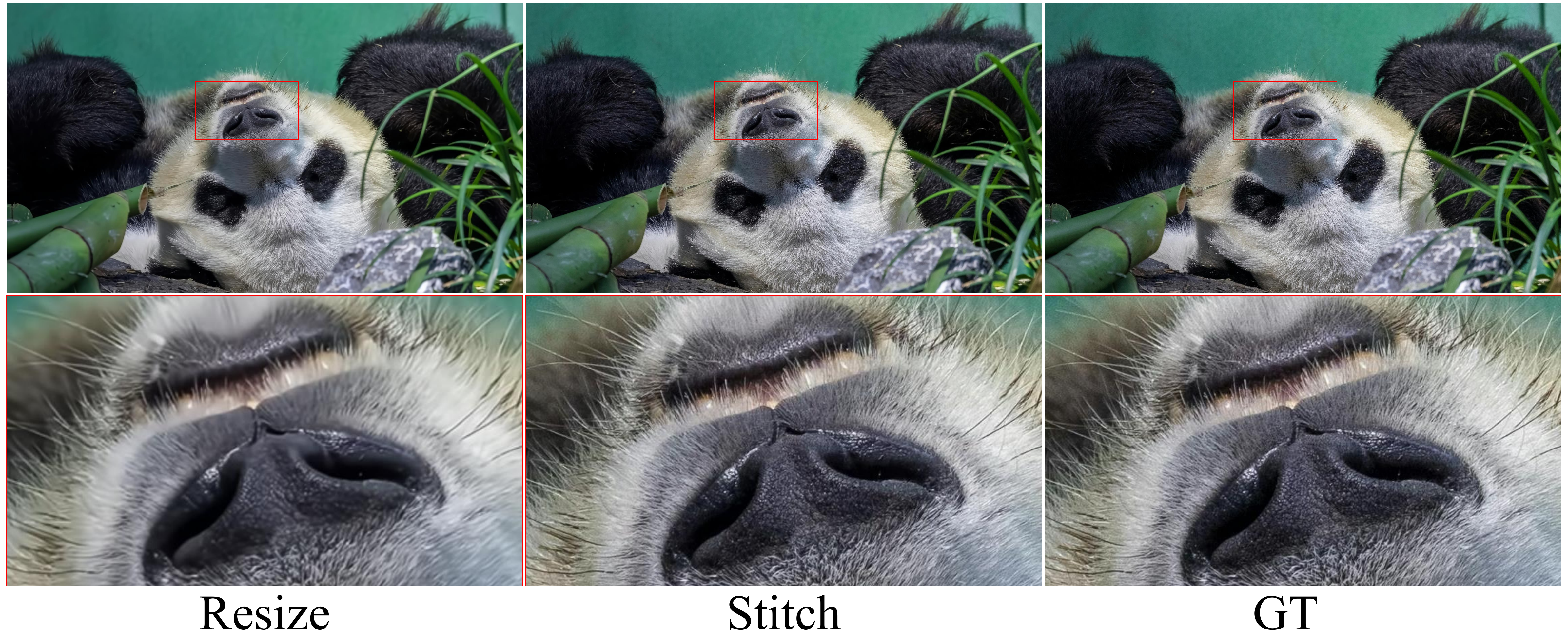}
     \vspace{-0.8em}
	\caption{A visualization comparison between the resize and stitch strategies for Restormer, revealing that the resize strategy introduces noticeable blurring artifacts due to the downsampling–upsampling operations. } 
	\label{fig:resize_split}
\end{figure}

\begin{table}[!t]
\caption{Comparison of SSIM and PSNR under resize and split strategies for Restormer and SwinIR at two noise levels.}
\centering
\resizebox{\linewidth}{!}{%
\begin{tabular}{c|c|c|c}
\hline

\textbf{Method} & \textbf{Strategy} & \textbf{Noise15} & \textbf{Noise25} \\
\hline
\multicolumn{1}{c|}{\multirow{2}{*}{\textbf{Restormer}}} 
& Resize & 0.8975 / 29.36 & 0.8767 / 28.24 \\
& Split  & 0.9606 / 34.49 & 0.9367 / 32.09 \\
\hline
\multicolumn{1}{c|}{\multirow{2}{*}{\textbf{SwinIR}}} 
& Resize & 0.8990 / 29.59 & 0.8747 / 28.20 \\
& Split  & 0.9602 / 34.47 & 0.9402 / 32.33 \\
\hline
\end{tabular} 
}

\label{tab:strategy_comparison}
\end{table}

In this section, we compare our method with SOTA approaches across five public UHD image restoration benchmarks, including low-light enhancement, deraining, deblurring, and dehazing, as well as moiré pattern removal. Based on our proposed LSUHDIR dataset, we establish two additional benchmarks: UHD image denoising (UHD-Noise) and JPEG compression artifact removal (UHD-JPEG). Extensive experiments are conducted on both tasks, demonstrating the effectiveness of our method. Furthermore, we extend IERR to general image restoration tasks, including general image dehazing, low-light image enhancement, and underwater image enhancement. Experimental results show that our method serves as a versatile framework for various image restoration scenarios.

\subsection{Experimental Settings}

\begin{table}[!t]
    \caption{\small{Quantitative comparison on the UHD-LL \cite{li2023embedding}.}}
     \vspace{-0.8em}
	\begin{center}
		\resizebox{8cm}{!}{
		\begin{tabular}{c|c|c|c|c}
			\hline
			\textbf{Methods} & PSNR$\uparrow$ & SSIM$\uparrow$ & LPIPS$\downarrow$   &Parameter$\downarrow$ \\
			\hline
            DiffLL \cite{jiang2023low} & 21.36 & 0.872 & 0.239 & 17.29M \\
            LLFormer \cite{wang2023ultra} & 22.79 & 0.853 & 0.264 & 13.15M \\
            UHDFour \cite{li2023embedding}  & 26.22 & 0.900 & 0.239 & 17.54M \\
			Wave-Mamba \cite{zou2024wave} & 27.35 & 0.913 & \textbf{0.185} & 1.258M \\
            UHDFormer \cite{wang2024correlation} & 27.11 & 0.927 & 0.245 & \textbf{0.339M} \\
	        UHDDIP \cite{wang2024ultra} & 26.74 & 0.928  & 0.207 & 0.81M \\
            D2Net \cite{wu2025dropout  } & 23.01 & 0.906 &0.286 & 5.22M \\
            DreamUHD \cite{ liu2025dreamuhd } & 27.72 & 0.928 & 0.205 & 1.215M \\
            UHD-processer \cite{ liu2025uhd } & 27.22 & 0.929 & 0.204 & 1.6M \\
            UHDPromer \cite{wang2026neural} & 27.15 & 0.928 & 0.211 & 0.743M \\
            \hline
            ERR  & 27.57 & 0.932 & 0.214 & 1.131M \\
                
            IERR  & \textbf{27.87} & \textbf{0.932} & 0.212 & 0.503M \\
			\hline
		\end{tabular}
		}
	\end{center}
	\label{table:ll}
\end{table}

\begin{table}[!t]
\vspace{-0.5em}
    \caption{\small{Quantitative comparison on the 4K-Rain13k \cite{chen2024towards}.}}
     \vspace{-0.8em}
	\begin{center}
		\resizebox{8cm}{!}{
		\begin{tabular}{c|c|c|c|c}
			\hline
			\textbf{Methods} & PSNR$\uparrow$ & SSIM$\uparrow$ & LPIPS$\downarrow$ & Parameter$\downarrow$ \\
			\hline
            SPDNet \cite{yi2021structure}  & 31.81 & 0.922 & 0.195 & 3.04M \\
            IDT \cite{xiao2022image} & 32.91 & 0.948 & 0.124 & 16.41M \\
            Restormer \cite{zamir2022restormer}& 33.02 & 0.933 & 0.173 & 26.12M \\
            DRSformer \cite{chen2023learning} & 32.94 & 0.933 & 0.171 & 33.65M \\
            UDRFormer \cite{chen2023learning}& 33.36 & 0.946 & 0.122 & 8.53M \\
	        UDR-Mixer \cite{chen2024towards} & 34.28 & 0.951 & 0.133 & 4.90M \\
            \hline
            ERR  & 34.48 & 0.952 & 0.120 &  1.131M \\

               IERR  & \textbf{34.89} & \textbf{0.955} & \textbf{0.111} &  \textbf{0.503M} \\
			\hline
		\end{tabular}
		}
	\end{center}
	\label{table:rain}
    \vspace{-0.5cm}
\end{table}

\begin{figure*}[t]
	\centering
	\includegraphics[width=0.95\linewidth]{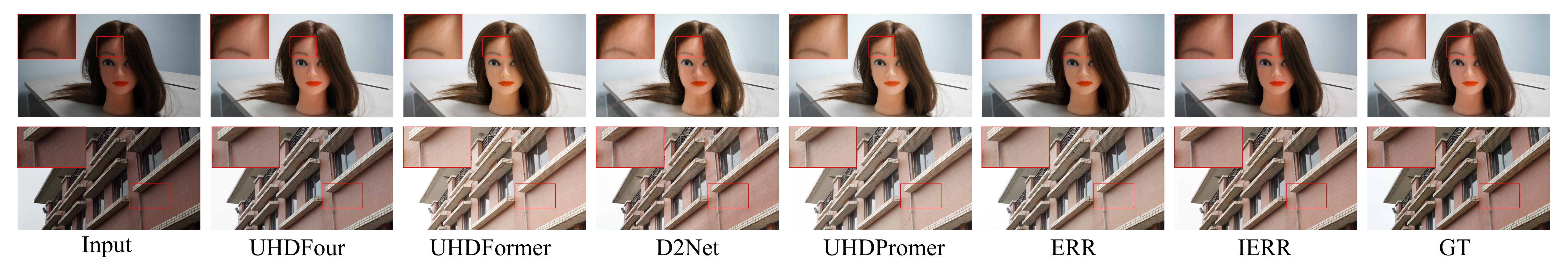}
	\caption{Visual comparison with other SOTA methods on the UHD-LL \cite{li2023embedding}. } 
	\label{fig:uhd_ll}
\end{figure*}

\noindent \textbf{Implementation details}. All experiments are conducted on an NVIDIA A6000 GPU. For the public ultra-high-definition (UHD) restoration dataset, we train the models using the AdamW optimizer with an initial learning rate of 0.0005, which is gradually reduced to 1e-7 after 100k iterations using cosine annealing \cite{loshchilov2016sgdr}. For our constructed UHD-Noise and UHD-JPEG datasets, we train for 500k iterations. All compared methods follow the same experimental settings to ensure a fair comparison. For general restoration tasks, given the lower resolution of the data, we adapt to these tasks by adjusting the downsampling factors and model parameters of ZFE and LFR.

\noindent \textbf{UHD IR datasets}. We evaluate our method on five UHD  restoration benchmarks. For low-light enhancement, we conduct experiments using the UHD-LL dataset \cite{li2023embedding}. The image deraining performance is assessed on the 4K-Rain13k dataset \cite{chen2024towards}. For the tasks of image dehazing and deblurring, we utilize the UHD-Haze \cite{wang2024correlation}  and UHD-Blur \cite{wang2024correlation}  datasets. For moiré pattern removal, we conduct experiments on the UHDM dataset \cite{yu2022towards}. In addition, we conduct denoising and JPEG compression artifact removal experiments on our proposed UHD-Noise and UHD-JPEG benchmarks.  

\noindent \textbf{General IR datasets}. We extend our method across five image restoration tasks. Specifically, we adopt LOL-v1 \cite{wei2018deep} and LOL-v2 \cite{yang2021sparse} for low-light image enhancement;  O-HAZE  \cite{ancuti2018haze}, NH-HAZE  \cite{ancuti2020nh}, and DENSE-HAZE  \cite{ancuti2019dense} for image dehazing; and UIEB  \cite{li2019underwater} and LSUI  \cite{PengZB23} for underwater image enhancement.

\begin{table}[!t]
    \caption{\small{Quantitative comparison on the UHD-Haze \cite{wang2024correlation}.}}
\vspace{-0.8em}
	\begin{center}
		\resizebox{7cm}{!}{
		\begin{tabular}{c|c|c|c}
			\hline
			\textbf{Methods} & PSNR$\uparrow$ & SSIM$\uparrow$ &Parameter$\downarrow$ \\
			\hline
            UHD  \cite{zheng2021ultra} & 18.04 & 0.811 & 34.5M \\
            DehazeFormer \cite{song2023vision} & 15.37 & 0.725 & 2.5M \\
            UHDDIP \cite{wang2024ultra} & 24.69 & 0.952 & 0.81M  \\
	        UHDFormer \cite{wang2024correlation} & 22.59 & 0.943 & \textbf{0.339M} \\
           DreamUHD \cite{liu2025dreamuhd} & 24.35& 0.945 & 1.215M \\
            D2Net \cite{wu2025dropout} & 24.88 & 0.944 & 5.22M \\
            UHD-processer \cite{ liu2025uhd } & 23.24 & 0.953 &  1.6M\\
            UHDPromer \cite{wang2026neural} & 22.72 & 0.943 & 0.743M \\
            \hline
            ERR  & 25.12 & 0.950 & 1.131M \\
                IERR  & \textbf{27.06} & \textbf{0.958} & 0.503M \\
			\hline
		\end{tabular}
		}
	\end{center}
	\label{table:haze}
\end{table}

\begin{table}[!t]
    \caption{\small{Quantitative comparison on the UHD-Blur \cite{wang2024correlation}.}}
   \vspace{-0.8em}
	\begin{center}
		\resizebox{7cm}{!}{
		\begin{tabular}{c|c|c|c}
			\hline
			\textbf{Methods} & PSNR$\uparrow$ & SSIM$\uparrow$ &Parameter$\downarrow$ \\
			\hline
            Restormer \cite{zamir2022restormer}& 25.21 & 0.693 & 26.1M \\
            Uformer \cite{wang2022uformer} & 25.27 & 0.737 & 20.6M \\
            FFTformer \cite{kong2023efficient}& 25.41 & 0.725 & 16.6M \\
             UHDDIP \cite{wang2024ultra} & 29.51 & 0.858 & 0.81M  \\
	        UHDFormer \cite{wang2024correlation} & 28.82 & 0.844 & \textbf{0.339M} \\
             DreamUHD \cite{liu2025dreamuhd} & 29.33 & 0.852 & 1.456M \\
             D2Net \cite{wu2025dropout} & 30.46 & 0.872 & 5.22M \\
             UHD-processer \cite{ liu2025uhd } & 28.91 & 0.851 &  1.6M \\
             UHDPromer \cite{wang2026neural} & 29.52 & 0.858 & 0.743M \\
            \hline
           ERR & 29.72 & 0.861 & 1.131M \\
           IERR & \textbf{30.53} & \textbf{0.873} & 0.503M \\
			\hline
		\end{tabular}
		}
	\end{center}
	\label{table:blur}
	\vspace{-1.3em}
\end{table}

\begin{table}[!t]
    \caption{\small{Quantitative comparison on the UHDM \cite{yu2022towards}.}}
 \vspace{-0.8em}
	\begin{center}
		\resizebox{7cm}{!}{
		\begin{tabular}{c|c|c|c}
			\hline
			\textbf{Methods} & PSNR$\uparrow$ & SSIM$\uparrow$ &Parameter$\downarrow$ \\
			\hline
            MBCNN \cite{ZhengYSL20} & 21.41  & 0.793 & 14.192M \\
            FHDe2Net \cite{HeWSD20}& 20.34 & 0.749 & 13.571M \\
             ESDNet \cite{yu2022towards} & 22.12 & 0.795  & 5.934M  \\
	        ESDNet-L \cite{yu2022towards} & 22.42 & 0.798  & 10.623M \\
            UHDFormer \cite{wang2024correlation} & 21.34 & 0.811  & \textbf{0.339M} \\
             
            \hline
           ERR & 22.77 & 0.821   & 1.131M \\
           IERR & \textbf{23.34} & \textbf{0.822} & 0.503M \\

			\hline
		\end{tabular}
		}
	\end{center}
	\label{table:moire}
	\vspace{-1.3em}
\end{table}

\begin{figure*}[t]
	\centering
	\includegraphics[width=0.8\linewidth]{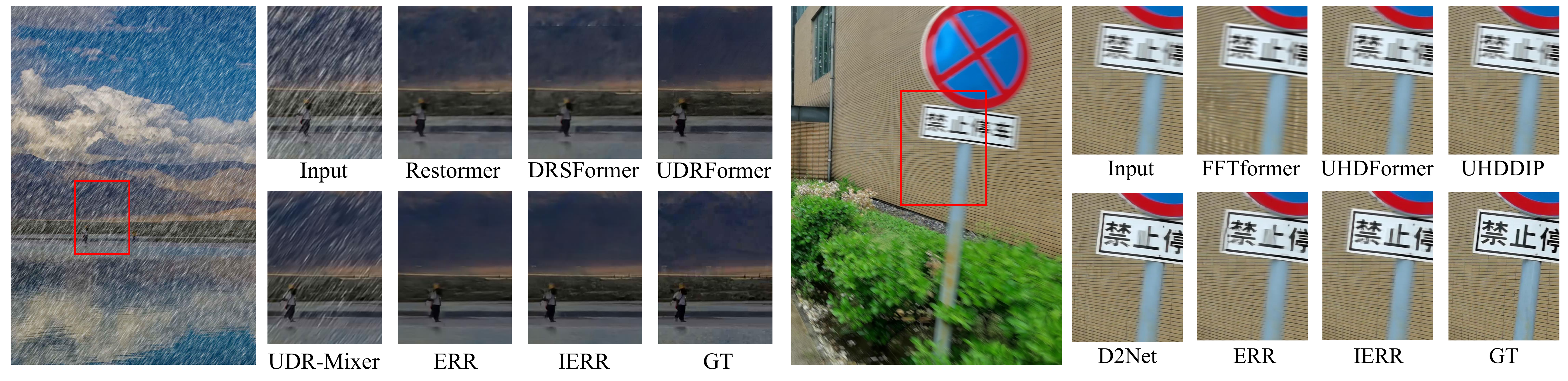}
	\caption{Visual comparison with other SOTA methods on the 4K-Rain13k \cite{chen2024towards} and UHD-Blur \cite{wang2024correlation}. } 
	\vspace{-0.1cm}
	\label{fig:uhd_blur}
\end{figure*}
\begin{figure*}[t]
	\centering
	\includegraphics[width=0.8 \linewidth]{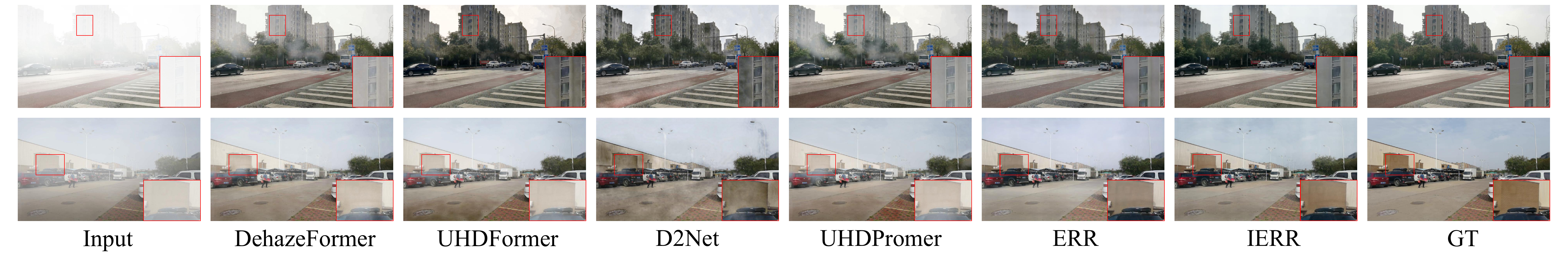}
	\caption{Visual comparison with other SOTA methods on the UHD-Haze \cite{wang2024correlation}. } 
	\label{fig:uhd_haze}
\end{figure*}

\begin{figure}[t]
	\centering
	\includegraphics[width=0.9 \linewidth]{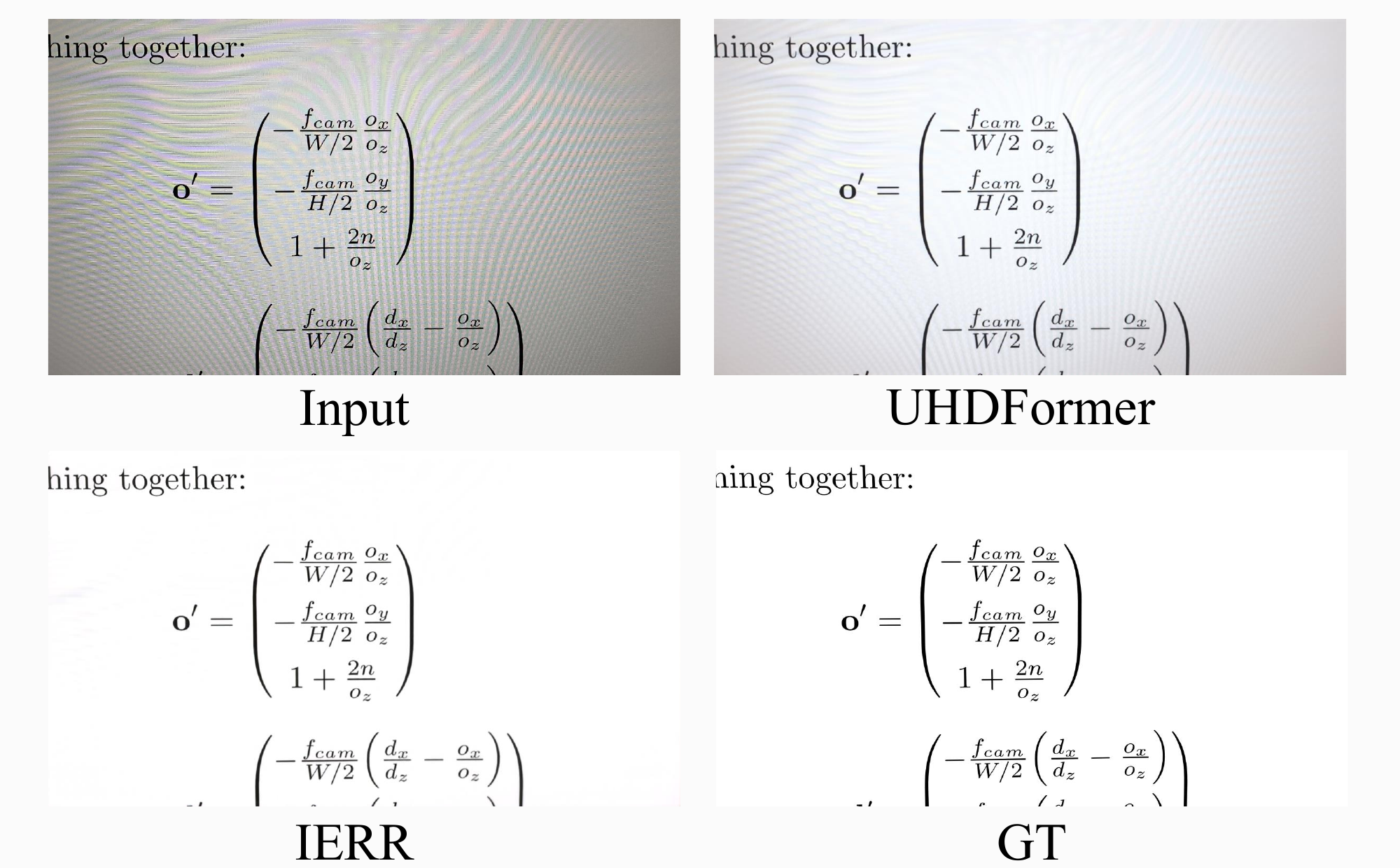}
     \vspace{-0.8em}
	\caption{Visual comparison on the UHDM  \cite{yu2022towards}.  } 
	\vspace{-0.1cm}
	\label{fig:UHDM}
\end{figure}
\begin{figure*}[t]
	\centering
	\includegraphics[width=0.85 \linewidth]{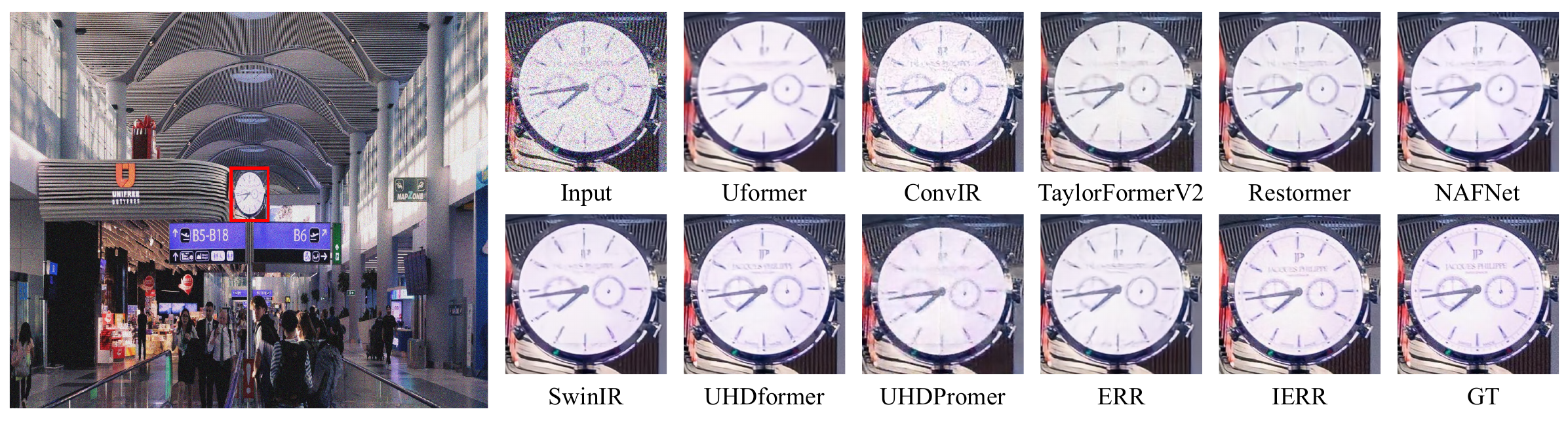}
	\caption{Visual comparison with other SOTA methods on our UHD-Noise. } 
	\vspace{-0.1cm}
	\label{fig:uhd_denoise}
\end{figure*}
\begin{figure*}[t]
	\centering
	\includegraphics[width=0.85 \linewidth]{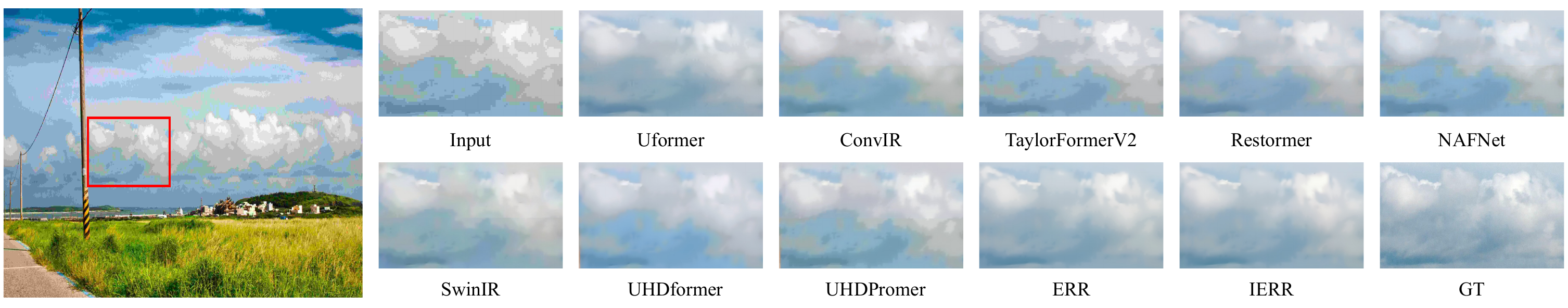}
	\caption{Visual comparison with other SOTA methods on our UHD-JPEG. } 
	\vspace{-0.1cm}
	\label{fig:uhd_dejpeg}
\end{figure*}

\begin{table*}[!t]
	\centering

	\caption{Quantitative comparison of different noise levels \(\sigma\) on our UHD-Noise for Gaussian denoising. The FLOPs are computed with an input size of 1024 × 1024.}
	\resizebox{0.95\width}{!}{
		\begin{tabular}{c c c c c c c c c c c c}
			\toprule
			\multirow{2}{*}{\textbf{Methods}} 
			& \multicolumn{3}{c}{\(\sigma = 15\)} 
			& \multicolumn{3}{c}{\(\sigma = 25\)} 
			& \multicolumn{3}{c}{\(\sigma = 50\)}      \\
			
			\cmidrule(lr){2-4}\cmidrule(lr){5-7}\cmidrule(lr){8-10}
			&	PSNR & SSIM & LPIPS & PSNR & SSIM  & LPIPS & PSNR &SSIM  & LPIPS &Params & FLOPs \\
			\midrule

			   SwinIR \cite{liang2021swinir}&  34.47 & 0.9602 & 0.0953 & 
                            32.33 & 0.9402 & 0.1473 & 
                            29.64 & 0.9031 & 0.2581 &
                            11.456M & 12.034T \\
                             
			   Restormer \cite{zamir2022restormer}& 33.82 & 0.9552 & \textbf{0.0890} &
                             31.93 & 0.9352 & 0.1509 &
                            29.34 & 0.8991 & 0.2561 &
                            26.097M & 2.256T \\
                  Uformer \cite{wang2022uformer}& 31.78 & 0.9309 & 0.2576 &
                           30.20 & 0.9080 & 0.3182 &
                           28.06 & 0.8721 & 0.3936 &
                           50.392M & 1.372T \\
                NAFNet \cite{chen2022simple}& 34.25 & 0.9583 & 0.1157 & 
                        32.09 & 0.9365 & 0.1850 &
                        29.41 & 0.9004 & 0.2617 & 
                        115.865M & 1.012T \\
                ConvIR \cite{cui2024revitalizing}& 33.12 & 0.9479 & 0.1041 &
                        31.02 & 0.9218 & 0.1602 &
                        28.26 & 0.8650 & 0.2479 & 
                        14.816M & 2.063T \\
                TaylorFormerV2 \cite{jin2025mb}& 33.21 & 0.9434 & 0.1334 & 
                                   28.63 & 0.8742 & 0.2950 &
                                   28.27 & 0.8705 & 0.2908 & 7.268M & 1.144T \\
			
			D2Net \cite{wu2025dropout} &34.89 & 0.9608 & 0.1090&32.38&0.9354&0.1867& 29.77  &0.9039   &0.2616  & 5.22M & 595.39G\\
            UHDformer \cite{wang2024correlation}& 34.68  & 0.9583  &0.1073 & 32.29 & 0.9369  &0.1610& 27.21 &0.8458 & 0.3769 &\textbf{0.339M}&48.37G\\
            UHDPromer \cite{wang2026neural}& 33.74  & 0.9533  & 0.1116  & 31.53 & 0.9295  &  0.1785& 28.58 & 0.8852 & 0.2853 & 0.743M& \textbf{32.56G} \\
			\midrule
            ERR     & 34.94 & 0.9627  &0.0974 & 32.49 & 0.9417  &0.1425  
                   & 29.55 & 0.9046  &0.2470 & 1.131M & 38.59G\\
			IERR     &   \textbf{35.07} &   \textbf{0.9629} &  0.0954&   \textbf{32.63} &  \textbf{0.9420}&	  \textbf{0.1385}
            & \textbf{29.81}&	  \textbf{0.9060  }&	  \textbf{0.2354} & 0.503M & 36.24G \\
			\bottomrule
	\end{tabular}  }

	\label{tab:denoise}
\end{table*}

\begin{table*}[!t]
	\centering

	\caption{Quantitative comparison of different noise levels $q$  on our UHD-JPEG  for JPEG compression artifact removal. The FLOPs are computed with an input size of 1024 × 1024.}
	\resizebox{0.95 \width}{!}{
		\begin{tabular}{c c c c c c c c c c c c}
			\toprule
			\multirow{2}{*}{\textbf{Methods}} 
			& \multicolumn{3}{c}{\(q = 5\)} 
			& \multicolumn{3}{c}{\(q = 10\)} 
			& \multicolumn{3}{c}{\(q = 30\)}      \\
			
			\cmidrule(lr){2-4}\cmidrule(lr){5-7}\cmidrule(lr){8-10}
			&	PSNR & SSIM & LPIPS & PSNR & SSIM  & LPIPS & PSNR &SSIM  & LPIPS &Params & FLOPs \\
			\midrule

			   SwinIR \cite{liang2021swinir}&   26.67 & 0.8084 & 0.3876 & 
                             29.33 & 0.8902 & 0.2943 &
                             33.84 & 0.9573 & 0.1146 &
                             11.456M & 12.034T \\
			   Restormer  \cite{zamir2022restormer}& 26.81 & 0.8171 & 0.3858 &
                            29.35 & 0.8941 & 0.2870 &
                            33.79 & 0.9573 & 0.1105 &
                            26.097M & 2.256T \\
                  Uformer \cite{wang2022uformer}& 26.69 & 0.8162 & 0.4018 &
                            28.86 & 0.8854 & 0.3330 &
                            32.32 & 0.9446 & 0.1860 &
                            50.392M & 1.372T \\
                NAFNet \cite{chen2022simple}&  26.42 & 0.8031 & 0.3931 & 
                            29.73 & 0.9003 & 0.2845 &
                            33.77 & 0.9569 & 0.1121 &
                            115.865M & 1.012T \\
                ConvIR \cite{cui2024revitalizing}& 26.55 & 0.8041 & 0.3856 &
                         29.31 & 0.8905 & 0.2872 &
                         33.98 & 0.9586 & 0.1097 &
                         14.816M & 2.063T \\
                TaylorFormerV2 \cite{jin2025mb}& 26.24 & 0.7973 & 0.3977 &
                                29.17 & 0.8888 & 0.3009 &
                                33.45 & 0.9551 & \textbf{0.1085} &
                                7.268M & 1.144T \\
			
			D2Net  \cite{wu2025dropout} & 26.95  &  0.8131   &   0.3976 &
              30.02 & 0.9009   &  0.3009 &  
                34.63 &   0.9600  &   0.1192&5.22M & 595.39G \\
            
            UHDformer \cite{wang2024correlation}& 26.72  & 0.8096  &   0.4111  & 
            29.81 & 0.8972 &     0.3033 & 
            34.49 & 0.9590  & 0.1219&\textbf{0.339M} & 48.37G \\  
            UHDPromer \cite{wang2026neural}& 26.37  &0.8021   &0.3917   & 29.31  & 0.8930  & 0.2920 & 33.71 & 0.9566  & 0.1148 & 0.743M&\textbf{32.56G}\\

			\midrule
            ERR     &  27.39 & 0.8415  &   0.3864
                  & 30.19  & 0.9064  & 0.2851
                 &  34.65 &0.9611  & 0.1090 &1.131M & 38.59G\\
			IERR     &   \textbf{27.56} &   \textbf{0.8427} &  \textbf{0.3841}&   \textbf{30.33} &  \textbf{0.9078}&	  \textbf{0.2839}& 
            \textbf{34.77}&	  \textbf{0.9609}&	  0.1120&0.503M & 36.24G \\
			\bottomrule
	\end{tabular}  }

	\label{tab:jpeg}
\end{table*}

\begin{table*}[!t]
\centering
\caption{Quantitative comparison on three low-light benchmarks.}
\resizebox{\linewidth}{!}{
\begin{tabular}{c c c c c c c c c c}
\toprule
 \multicolumn{2}{c}{\textbf{Methods}} 
& Restormer \cite{zamir2022restormer} & LEDNet\cite{zhou2022lednet} & SNR-Net\cite{xu2022snr} & LLFormer \cite{wang2023ultra}
& RetinexFormer \cite{cai2023retinexformer} & SFHformer\cite{jiang2024fast} & CIDNet \cite{yan2025hvi} & \textbf{IERR} \\
\midrule
\multirow{2}{*}{LOL-v1 \cite{wei2018deep}} 
& PSNR↑ & 22.37 & 20.63 & 24.61 & 23.65 & 25.15 & 24.29 & 23.50 & \textbf{25.18} \\
& SSIM↑ & 0.816 & 0.823 & 0.842 & 0.816 & 0.846 & 0.862 & 0.870 & \textbf{0.870} \\
\midrule
\multirow{2}{*}{LOL-v2-real \cite{yang2021sparse}} 
& PSNR$\uparrow$ & 18.69 & 19.94 & 21.48 & 20.06 & 22.79 & 23.78 & 23.43 & \textbf{23.80} \\
& SSIM$\uparrow$ & 0.834 & 0.827 & 0.849 & 0.792 & 0.840 & \textbf{0.872} & 0.862 & 0.855 \\
\midrule
\multirow{2}{*}{LOL-v2-syn \cite{yang2021sparse}} 
& PSNR$\uparrow$ & 21.41 & 23.71 & 24.14 & 24.04 & 25.67 & 25.80 & 25.71 & \textbf{26.21} \\
& SSIM$\uparrow$ & 0.830 & 0.914 & 0.928 & 0.909 & 0.930 & 0.937 & 0.942 & \textbf{0.943} \\
\midrule
overhead &Params$\downarrow$
& 26.10M & 7.07M & 4.01M & 24.55M & 1.61M & \textbf{1.04M} & 1.88M & 1.42M \\
\bottomrule
\end{tabular}
}
\label{tab:lol}
\end{table*}

\begin{table*}[!t]
\centering
\caption{Quantitative comparison on three dehazing benchmarks.}
\resizebox{\linewidth}{!}{
\begin{tabular}{c c c c c c c c c c}
\toprule
 \multicolumn{2}{c}{\textbf{Methods}} 
& MSBDN  \cite{dong2020multi} & FFA-Net \cite{qin2020ffa} & Dehamer \cite{guo2022image} & C2PNet \cite{zheng2023curricular}
& ConvIR \cite{cui2024revitalizing} & TaylorFormer \cite{qiu2023mb} & TaylorFormerV2 \cite{jin2025mb} & \textbf{IERR} \\
\midrule
\multirow{2}{*}{O-HAZE \cite{ancuti2018haze}} 
& PSNR↑ & 24.36 & 22.12 & 25.11 & 25.20 & 25.36 & 25.31 & 25.29 & \textbf{25.45} \\
& SSIM↑ & 0.749 & 0.770 & 0.777 & 0.785 & 0.780 & 0.782 & 0.790 & \textbf{0.790} \\
\midrule
\multirow{2}{*}{Dense-Haze \cite{ancuti2019dense} } 
& PSNR$\uparrow$ & 15.13 & 15.70 & 16.62 & 16.88 & 16.86 & 16.44 & 16.95 & \textbf{17.23} \\
& SSIM$\uparrow$ & 0.555 & 0.549 & 0.560 & 0.573 & 0.600 & 0.566 & 0.621 & \textbf{0.627} \\
\midrule
\multirow{2}{*}{NH-HAZE \cite{ancuti2020nh}} 
& PSNR$\uparrow$ & 17.97 & 18.13 & 20.66 & 20.24 & 20.66 & 20.49 & 20.73 & \textbf{21.21} \\
& SSIM$\uparrow$ & 0.659 & 0.647 & 0.684 & 0.687 & 0.691 & 0.692 & 0.703 & \textbf{0.703} \\
\midrule
overhead &Params$\downarrow$
& 31.35M & 4.46M  &  132.50M  & 7.17M & 8.63M & 7.43M & 2.63M & \textbf{0.503M} \\
\bottomrule
\end{tabular}
}
\label{tab:ohaze}
\end{table*}
\begin{table}[!t]
	\centering

	\caption{Quantitative comparison for underwater image enhancement.}
	\resizebox{1\width}{!}{
		\begin{tabular}{c  c c c c c }
			\toprule
			\multirow{2}{*}{\textbf{Methods}} 
			& \multicolumn{2}{c}{UIEB \cite{li2019underwater} } & \multicolumn{2}{c}{LSUI \cite{PengZB23}} 
			   \\
			
			\cmidrule(lr){2-3}   \cmidrule(lr){4-5}
			&	PSNR & SSIM    &	PSNR & SSIM & Params \\
			\midrule

                  U-shape \cite{PengZB23} & 16.01 & 0.8180 & 
                           23.26 & 0.8241 &65.6M \\
                DM-water \cite{TangKI23}&21.79 & 0.8517 & 
                        24.15 & 0.8716 & - \\
                CECF \cite{CongGH24}& 21.35 & 0.7748  & 
       26.12 & 0.8664 & - \\
               HCLR \cite{ZhouSLJZLZF24}& 22.24  & 0.9002 
                                   &26.64& \textbf{0.8815}& \textbf{4.87M} \\
               MambaIR \cite{guo2025mambair} & 21.00 & 0.8618
                                    &24.90 & 0.8771 & 16.7M \\
			
			\midrule
         
			IERR     &   \textbf{24.25} & \textbf{0.9318}  &   \textbf{27.54} & 0.8871  &5.13M     \\
			\bottomrule
	\end{tabular}  }

	\label{tab:underwater}
\end{table}

\noindent \textbf{Evaluation}. We mainly adopt peak signal to noise
ratio (PSNR) \cite{hore2010image}  and structural similarity (SSIM) \cite{wang2004image}  to evaluate the performance
of methods. Additionally, LPIPS \cite{zhang2018unreasonable} is utilized to evaluate 
perceptual performance. Currently, for general IR methods that are incapable of 4K full-resolution inference, there are two common strategies \cite{li2023embedding}: (1) resize — downsample the input to the maximum resolution the model can handle, perform inference, and then upsample the output back to the original resolution; (2) stitch — divide the input into multiple non-overlapping patches, process each patch independently, and then stitch the results together to form the final output. For consistency, we apply a 2× downsampling in the resize strategy and divide the input into 4 non-overlapping patches for the stitch strategy. We employ Restormer and SwinIR as baselines to evaluate the effectiveness of these two strategies. As shown in Table \ref{tab:strategy_comparison}, the resize strategy yields lower PSNR and SSIM scores due to the information loss introduced by the downsampling and upsampling operations. This observation is further corroborated by the visual comparisons in Figure \ref{fig:resize_split}. Although the stitch strategy incurs higher computational overhead during inference, we adopt it across all experiments on our benchmark to ensure fair evaluation.

\subsection{Comparisons Results with UHD IR}

\subsubsection{UHD Low-light Image Enhancement}

We trained the proposed ERR and IERR models on the UHD-LL dataset and compared it against recent low-light enhancement methods, including 
DiffLL \cite{jiang2023low}, LLFormer \cite{wang2023ultra}, UHDFour \cite{li2023embedding}, UHDFormer \cite{wang2024correlation}, Wave-Mamba \cite{zou2024wave},  UHDDIP \cite{wang2024ultra},  D2Net \cite{wu2025dropout  }, DreamUHD \cite{liu2025dreamuhd}, UHD-processer \cite{ liu2025uhd}, and UHDPromer \cite{wang2026neural}. The quantitative results in Table \ref{table:ll} demonstrate that our ERR and IERR significantly enhance performance, improving PSNR and SSIM metrics and surpassing all baselines. Figure \ref{fig:uhd_ll} provides visual evaluations on the UHD-LL dataset, where our IERR preserves finer details and achieves superior perceptual quality.

\subsubsection{UHD Image Deraining}

We evaluate the effectiveness of UHD deraining on the 4K-Rain13k, comparing our ERR and IERR with recent methods, including 
SPDNet \cite{yi2021structure}, IDT \cite{xiao2022image}, Restormer \cite{zamir2022restormer}, UDRFormer \cite{chen2023learning}, and UDR-Mixer \cite{chen2024towards}. Quantitative results in Table \ref{table:rain} demonstrate that our method achieves the highest scores, while reducing model parameters. The visual results in Figure \ref{fig:uhd_blur} show that our method effectively removes rain streaks while preserving rich texture details.

\subsubsection{UHD Image Dehazing}

Table \ref{table:haze} summarizes the quantitative  results on the UHD-Haze, comparing our methods with recent  methods such as UHD \cite{zheng2021ultra}, 
DehazeFormer \cite{song2023vision},  UHDFormer \cite{wang2024correlation}, D2Net \cite{wu2025dropout  }, DreamUHD \cite{liu2025dreamuhd}, UHD-processer \cite{ liu2025uhd}, and UHDPromer \cite{wang2026neural}. Our IERR achieves a PSNR improvement of 1.94dB with a relatively small number of parameters, and attains the highest SSIM score. Qualitative results in  Figure \ref{fig:uhd_haze} demonstrate that our IERR effectively restores clear images, whereas other methods struggle to remove dense haze.

\subsubsection{UHD Image Deblurring}

We evaluate the deblurring performance of our model on the UHD-Blur dataset, comparing it with recent methods including 
Restormer \cite{zamir2022restormer}, Uformer \cite{wang2022uformer}, 
FFTformer \cite{kong2023efficient}, UHDFormer \cite{wang2024correlation}, D2Net \cite{wu2025dropout  }, DreamUHD \cite{liu2025dreamuhd}, UHD-processer \cite{ liu2025uhd}, and UHDPromer \cite{wang2026neural}. As shown in Table \ref{table:blur}, our approach achieves the highest PSNR and SSIM scores, demonstrating its superior effectiveness. Qualitative results in Figure \ref{fig:uhd_blur} reveal that our IERR generates clearer details.

\subsubsection{UHD Image Demoiréing}
We evaluate the demoiréing performance of our models on the UHDM dataset, comparing it with recent methods including 
MBCNN \cite{ZhengYSL20}, FHDe2Net \cite{HeWSD20}, ESDNet \cite{yu2022towards}, and UHDFormer  \cite{wang2024correlation}. 
As shown in Table \ref{table:moire}, our IERR achieves the highest PSNR and SSIM scores, demonstrating its superior effectiveness. Qualitative results in Figure \ref{fig:UHDM}  further reveal that IERR generates clearer details.

\subsubsection{UHD Image Denoising and JPEG Removal}
We conduct UHD denoising and JPEG compression artifact removal experiments on our proposed UHD-Noise and UHD-JPEG benchmarks, which are newly constructed in this work to evaluate restoration under realistic noise and compression degradations. For both tasks, we compare our approach with recent strong baselines including SwinIR \cite{liang2021swinir}, Restormer \cite{zamir2022restormer}, Uformer \cite{wang2022uformer}, NAFNet \cite{chen2022simple}, ConvIR \cite{cui2024revitalizing}, TaylorFormerV2 \cite{jin2025mb}, D2Net \cite{wu2025dropout} UHDFormer \cite{wang2024correlation}, and UHDPromer \cite{wang2026neural}.
As shown in Table \ref{tab:denoise}, our IERR achieves the best PSNR and SSIM scores on UHD-Noise, clearly surpassing existing methods. Qualitative comparisons in Figure \ref{fig:uhd_denoise} demonstrate that our IERR effectively suppresses noise and recovers clean textures with sharp structural details. 
For JPEG artifact removal, Table \ref{tab:jpeg} reports that IERR consistently outperforms competing approaches on UHD-JPEG, delivering significant gains in both PSNR and SSIM. Visual results in Figure \ref{fig:uhd_dejpeg} show that ERR and IERR reduce blocking artifacts and restore more natural image structures, whereas other methods leave noticeable compression traces.

\subsection{Comparisons Results with IR}
We adapt to various general image restoration tasks by adjusting the downsampling factors of ZFE and LFR. Extensive experimental results demonstrate that our 'from zero to detail' decoupling framework is not merely an algorithm designed specifically for UHD scenarios, but rather a general-purpose paradigm for image restoration. We primarily evaluate the effectiveness of IERR for image restoration. \textbf{More results can be found in the supplementary material}.

\subsubsection{Low-light Image Enhancement}
We evaluate the effectiveness of low-light image enhancement on the LOL-v1 \cite{wei2018deep} and LOL-v2 \cite{yang2021sparse} datasets, comparing our IERR with recent methods, including Restormer \cite{zamir2022restormer}, LEDNet\cite{zhou2022lednet}, SNR-Net\cite{xu2022snr}, LLFormer \cite{wang2023ultra}, RetinexFormer \cite{cai2023retinexformer}, SFHformer\cite{jiang2024fast}, and CIDNet \cite{yan2025hvi}. Due to the low image resolution of the LOL dataset, we adjust the downsampling factor of ZFE to 4 and disable downsampling in LFR for this experiment.  As shown in Table \ref{tab:lol}, our method outperforms others and yields the best PSNR scores. Notably, we do not employ the GT Mean strategy for all methods. 

\subsubsection{Image Dehazing}
We evaluate the effectiveness of image dehazing on the O-HAZE \cite{ancuti2018haze}, Dense-Haze \cite{ancuti2019dense}, and NH-HAZE \cite{ancuti2020nh},  comparing our IERR with recent methods, including JMSBDN  \cite{dong2020multi}, FFA-Net \cite{qin2020ffa}, Dehamer \cite{guo2022image}, C2PNet \cite{zheng2023curricular}, ConvIR \cite{cui2024revitalizing}, TaylorFormer \cite{qiu2023mb}, and TaylorFormerV2 \cite{jin2025mb}. Due to the fact that images in these datasets are all larger than 1K, we use the original IERR for this experiment. As shown in Table \ref{tab:ohaze}, our method yields the best PSNR using minimal parameters.

\subsubsection{Underwater Image Enhancement}
We evaluate the effectiveness of underwater image enhancement on the UIEB \cite{li2019underwater} and LSUI  \cite{PengZB23} datasets, comparing our IERR with recent methods, including  U-shape \cite{PengZB23}, DM-water \cite{TangKI23}, CECF \cite{CongGH24}, HCLR \cite{ZhouSLJZLZF24} and  MambaIR \cite{guo2025mambair}. Due to the low resolution of the UIEB and  LSUI dataset, we adjust the downsampling factor of ZFE to 4 and disable downsampling in LFR for this experiment. Quantitative results in Table \ref{tab:underwater} demonstrate that our method achieves the highest scores across all metrics. 

\begin{figure}[t]
	\centering
	\includegraphics[width=0.98\linewidth]{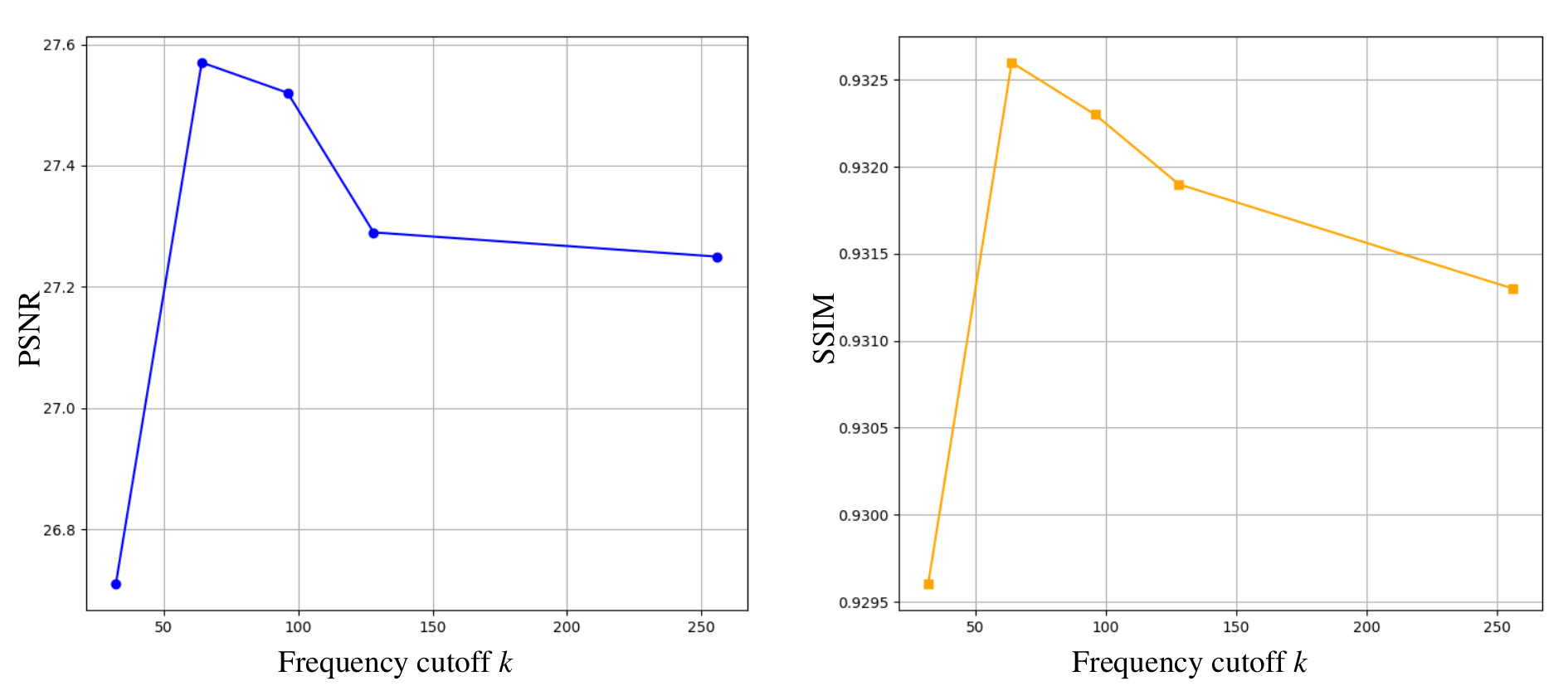}
     \vspace{-0.8em}
	\caption{Effect of different frequency cutoffs $k$.  } 
	\vspace{-0.1cm}
	\label{fig:cutoff}
\end{figure}

\begin{table}[!t]
\caption{Ablation study with different architecture on our ERR. PR refers to progressive residual, as indicated by the yellow arrows between each stage in Figure \ref{fig:3}.}
\centering
\label{tab:freq}

\setlength{\tabcolsep}{3pt} 
\renewcommand{\arraystretch}{1} 
\footnotesize 

\begin{tabular}{cccccccccc}
\specialrule{1.2pt}{0.2pt}{1pt}

Method  & $\mathcal{L}_{zf}$ & $\mathcal{L}_{lf}$  & $\mathcal{L}_{hf}$  & ZFE & LFR & HFR & PR & PSNR$\uparrow$ & SSIM$\uparrow$       \\\midrule
\midrule
A &  $\times$ & $\checkmark$ & $\checkmark$ & $\checkmark$ & $\checkmark$& $\checkmark$  & $\checkmark$  & 26.73 & 0.9286 \\
B &  $\checkmark$&  $\times$ & $\checkmark$ & $\checkmark$ & $\checkmark$& $\checkmark$  & $\checkmark$  & 26.57 & 0.9277 \\
C & $\checkmark$& $\checkmark$ & $\times$ & $\checkmark$ & $\checkmark$& $\checkmark$  & $\checkmark$ & 26.80  & 0.9290\\
\midrule
D &  $\times$ & $\checkmark$ & $\checkmark$ & $\times$ & $\checkmark$& $\checkmark$  & $\checkmark$ &27.01  & 0.9311 \\
E & $\checkmark$&  $\times$ & $\checkmark$ & $\checkmark$ & $\times$ & $\checkmark$  & $\checkmark$  & 26.81 & 0.9186 \\ 
F & $\checkmark$& $\checkmark$ & $\times$ & $\checkmark$ & $\checkmark$& $\times$   & $\checkmark$ & 27.04  & 0.9266 \\
\midrule
G & $\checkmark$& $\checkmark$ & $\checkmark$ & $\checkmark$ & $\checkmark$& $\checkmark$   & $\times$ &  26.99 & 0.9307
  \\
\midrule
\midrule
H & $\checkmark$& $\checkmark$ & $\checkmark$ & $\checkmark$ & $\checkmark$&$\checkmark$&$\checkmark$ & \textbf{27.57} & \textbf{0.9326 } \\
\specialrule{1.2pt}{0.2pt}{1pt}

\end{tabular}
 \vspace{-0.5em}
\label{tab:framerwork}
\end{table}

\begin{table}[t]
	\caption{Ablation study with the detail design of the ZFE. }
	\centering
	 \vspace{-1.2em}
	\resizebox{1\linewidth}{!}{
		\begin{tabular}{c c c  c c c c}
			\specialrule{1.2pt}{0.2pt}{1pt}
			\multicolumn{2}{c}{AAP} &\multicolumn{3}{c}{BBGM} &  \multicolumn{2}{c}{Metrics~} \\
			\cmidrule(lr){1-2}
			\cmidrule(lr){3-5}
            \cmidrule(lr){6-7}
			 $AvP_{1,1}$ & $AvP_{\frac{H}{8}, \frac{W}{8}}(x)$ & Fusion & Gate & Interaction & PSNR$\uparrow$ & SSIM$\uparrow$ \\
			\midrule
			\midrule
			$\times$  &$\times$ & $\times$ &$\times$ &$\times$ & 26.70  & 0.9289\\
			$\times$& $\checkmark$ & $\checkmark$& $\checkmark$ &$\checkmark$ & 27.16 & 0.9315\\
            $\checkmark$& $\times$ & $\checkmark$& $\checkmark$ &$\checkmark$ & 27.18 & 0.9316  \\
            \midrule

			$\checkmark$ & $\checkmark$ & $\times$& $\checkmark$   & $\checkmark$ & 26.00& 0.9251 \\
			$\checkmark$ & $\checkmark$ & $\checkmark$ & $\times$&$\checkmark$ & 26.20 &0.9268 \\
            $\checkmark$ & $\checkmark$ & $\checkmark$ & $\checkmark$&$\times$ & 26.04 &0.9256 \\
			\midrule
            \midrule
			$\checkmark$ & $\checkmark$ & $\checkmark$ &  $\checkmark$&$\checkmark$ & \textbf{27.57} & \textbf{0.9326 }\\
			\specialrule{1.2pt}{0.2pt}{1pt}
	\end{tabular}}
	\label{tab:zfe}
	 \vspace{-0.8 em}
\end{table}

\begin{figure}[!t]
	\centering
	\includegraphics[width=0.98\linewidth]{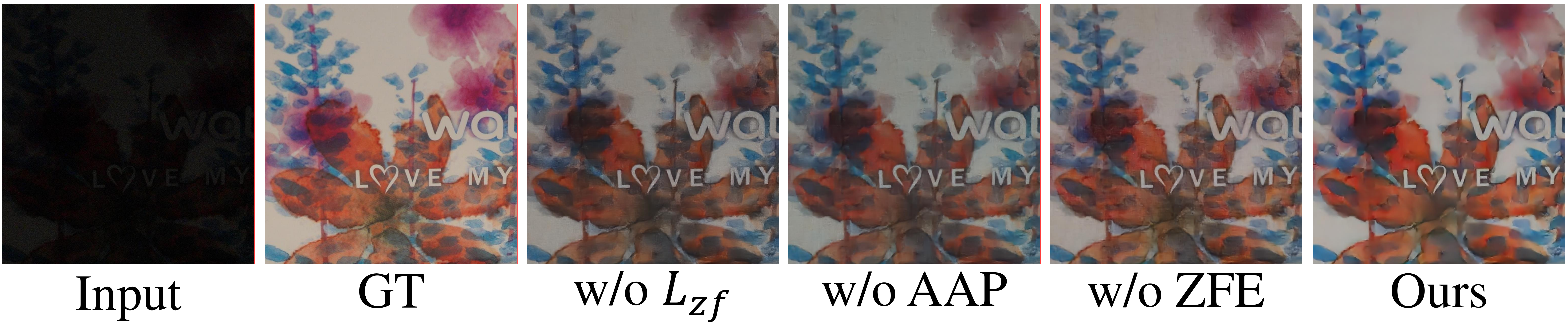}
    \vspace{-0.1 cm}
	\caption{Visual ablation results for the zero-frequency part. } 
	\vspace{-0.1 cm}
	\label{fig:ablow}
\end{figure}

\begin{figure}[t]
	\centering
	\includegraphics[width=0.98\linewidth]{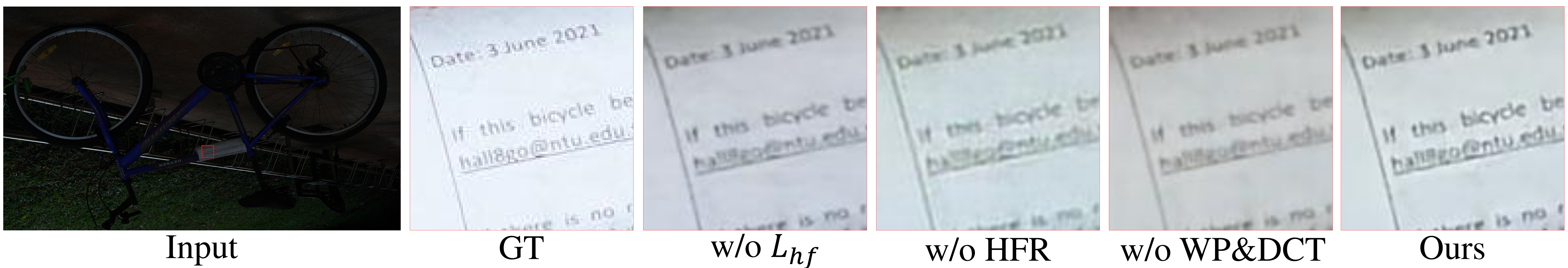}

        \vspace{-0.1 cm}
	\caption{Visual ablation results for the high-frequency part. } 
	\vspace{-0.1 cm}
	\label{fig:abhigh}
\end{figure}
\begin{figure}[t]
	\centering
	\includegraphics[width=0.98\linewidth]{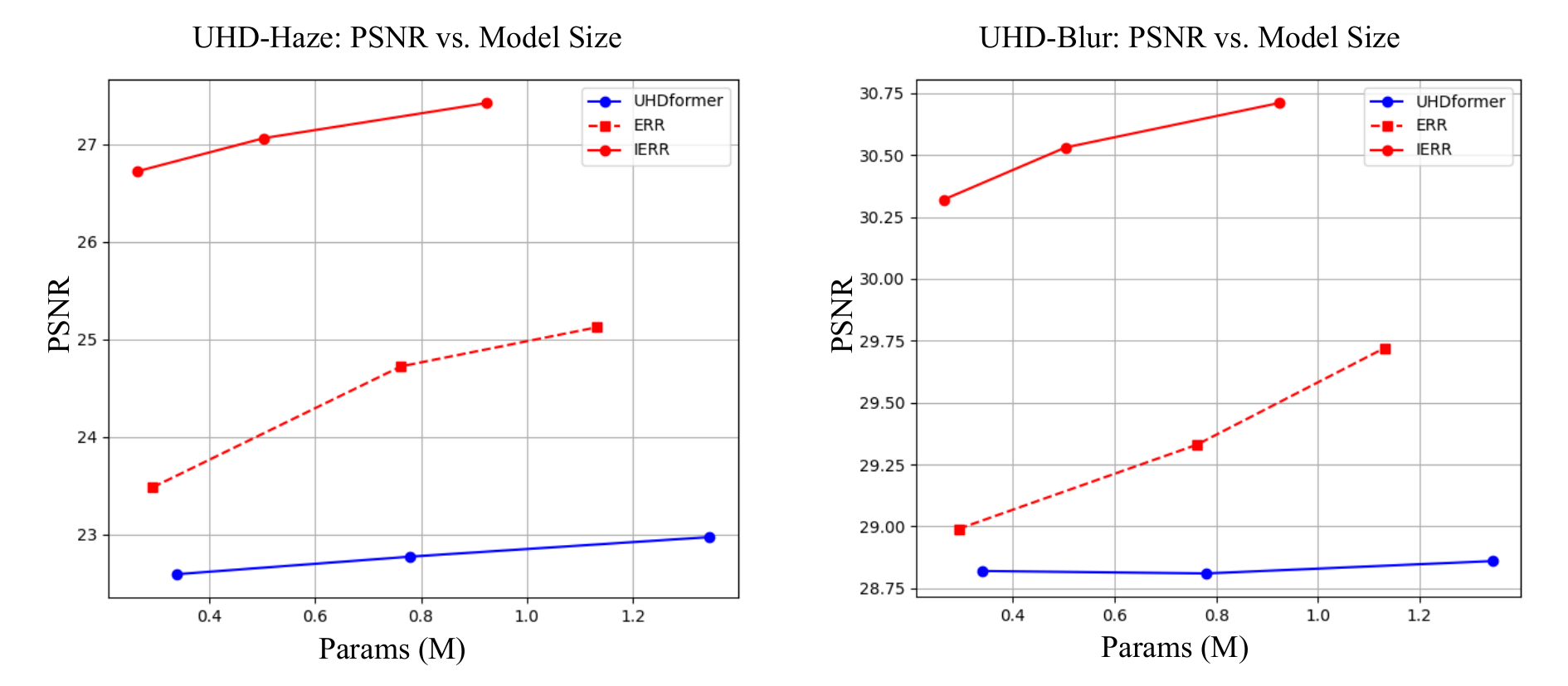}
     \vspace{-0.5em}
	\caption{Analysis of the model complexity.  } 
	\vspace{-0.1cm}
	\label{fig:complex}
\end{figure}

\begin{table}[!t]
    \caption{ Ablation study with the detail design of the HFR.}
	\centering
 \vspace{-0.6em}
	\resizebox{0.9\linewidth}{!}{
  {\setlength{\tabcolsep}{3pt}
		\begin{tabular}{c c c c  c  }
			\specialrule{1.2pt}{0.2pt}{1pt}
			
			Model & w/o DCT &  w/o WP&  w/o WP \& DCT  & ERR \\
			\midrule
			PSNR $\uparrow$&  26.88    &   26.97  &   25.78  &   \textbf{27.57}   \\
			SSIM $\uparrow$&   0.9308  &    0.9296   &  0.9247 & \textbf{0.9326 }     \\
			
			\specialrule{1.2pt}{0.2pt}{1pt}
	\end{tabular}}
    }
    \vspace{-1.0em}
	\label{tab:hfr}
	
\end{table}

\begin{table}[!t]
    \caption{ Ablation for FW-KAN.}
        \vspace{-0.6em}
	\centering

	\resizebox{1\linewidth}{!}{
  {\setlength{\tabcolsep}{4pt}
		\begin{tabular}{c c c c c  c  }
			\specialrule{1.2pt}{0.2pt}{1pt}

			Model & MLP-6 &  MLP-12 &  MLP-24 &  KAN  & FW-KAN \\
			\midrule
			PSNR $\uparrow$&  26.03 &   26.58  &  26.97 & 20.48  &    \textbf{27.57}   \\
			SSIM $\uparrow$&   0.9294 &    0.9297   &0.9311  & 0.8836  &  \textbf{0.9326 }     \\
            Parameter$\downarrow$ &   30.73K   &    35.63K & 45.42K & 27.57K &  \textbf{27.47}K   \\
			
			\specialrule{1.2pt}{0.2pt}{1pt}
	\end{tabular}}
    }
    \vspace{-1.1em}
	\label{tab:kan}
\end{table}

\begin{table}[!t]

	\caption{Ablation study for the two-stage variants on UHD-LL.}
	\vspace{-0.8em}
	\centering

	\resizebox{1\linewidth}{!}{
		{\setlength{\tabcolsep}{3pt}
			\begin{tabular}{c| c c  c | c   }
				\specialrule{1.2pt}{0.2pt}{1pt}
	
				Model & V1& V2& V3  &ERR \\
				\midrule
				PSNR/SSIM$\uparrow$ & 26.97/0.9221 &  27.15/0.9210 &  27.06/0.9190 &    \textbf{27.57/0.9326 }   \\
					\midrule
				 Params(M)$\downarrow$ &   1.131 & 1.50  &  1.149 &   1.131       \\
				 	FLOPs(G)$\downarrow$ & 307.52  &  738.98   &  252.59  &   307.52    \\

				\specialrule{1.2pt}{0.2pt}{1pt}
		\end{tabular}}
	}
	\vspace{-0.2 em}
	\label{tab:two_stage}
	
\end{table}

\begin{table}[t]
\caption{Ablation study with architecture on our IERR. }
\centering
\label{tab:freq}


\begin{tabular}{cccccccc}
\specialrule{1.2pt}{0.2pt}{1pt}

Method  & GPP & Down & $\mathcal{L}_{g}$   & LEM & ZR & PSNR$\uparrow$ & SSIM$\uparrow$       \\
\midrule
I &   $\checkmark$ & $\times$ &   $\times$  & $\times$  & $\times$  & 27.64 & 0.931 \\
II &  $\checkmark$&  $\checkmark$ &  $\times$  & $\times$  & $\times$   & 27.63 &0.932  \\
III & $\checkmark$& $\checkmark$ & $\checkmark$ &  $\times$  & $\times$  &   27.71& 0.931 \\
\midrule
IV &  $\checkmark$ & $\checkmark$ & $\checkmark$ & $\checkmark$   & $\times$  & 27.81 & 0.932 \\
\midrule
IERR& $\checkmark$& $\checkmark$ &$\checkmark$&$\checkmark$&$\checkmark$ & \textbf{27.87} & \textbf{0.932} \\
\specialrule{1.2pt}{0.2pt}{1pt}

\end{tabular}
 \vspace{-0.5em}
\label{tab:IERR}
\end{table}

\subsection{Ablation Study}

We perform comprehensive ablation experiments to verify the effectiveness of each contribution and design. All ablations are conducted on UHD-LL \cite{li2023embedding}.
\subsubsection{Ablation study for ERR framework}
In this section, we use the ERR model from our original conference version as the baseline and conduct an in-depth analysis of the proposed progressive spectral decoupling paradigm from multiple perspectives.

\noindent \textbf{Study of different frequency cutoffs $k$}.
   The ablation results with different cutoff values $k$, as shown in Figure~\ref{fig:cutoff}, demonstrate that both excessively small and large values of $k$ degrade performance. A small $k$ emphasizes high-frequency components, increasing the learning difficulty for the HFR branch, whereas a large $k$ introduces excessive low-frequency information, which hinders the LFR from effectively restoring content.
   

\noindent \textbf{Analysis with different architecture}. Table \ref{tab:framerwork} demonstrates the impact of all components on our ERR. Models A, B, and C validate the effectiveness of our frequency regularizations. 
When ZFE is omitted, the model (D) is a two-stage network with LFR and HFR. Similarly, when LFR or HFR is omitted, corresponding architectures are used. This ablation of models D, E, and F, aims to validate each stage’s effectiveness. The superior performance of D, E, and F over A, B, and C suggests that simply increasing the network without frequency constraints leads to challenges in optimization, thereby affecting the model’s performance. 
Furthermore, the most significant performance drop in Model E establishes LFR as the fundamental structural backbone, while Models D and F demonstrate that ZFE and HFR act as indispensable synergistic modules for ensuring global illumination consistency and high-fidelity textures, respectively.
Finally, when PR is omitted, we use the input as the residual at each stage. Model H outperforms G, indicating that the PR is more effective than using the input at each stage.

\noindent \textbf{Ablation with the ZFE}. In Table \ref{tab:zfe}, we investigate the impact of ZFE by focusing on the AAP unit for global prior generation and the BBGM for global prior integration. We first remove both AAP and BBGM, resulting in a pure Transformer at this stage, which yields significantly degraded performance—highlighting the essential role of these components. We then ablate the global and local $AvP$ branches within the AAP unit individually, and the results further validate the effectiveness of our prior generation design. Finally, removing individual components within BBGM causes substantial performance degradation, underscoring the critical importance of the prior integration strategy at this stage. Figure \ref{fig:ablow} provides visual comparisons of the zero-frequency ablation results.

\noindent \textbf{Ablation with the HFR}. Table \ref{tab:hfr} demonstrates the effectiveness of the DCT and WR operations in the HFR module. Removing either DCT or WR leads to noticeable performance drops, indicating that window-based partitioning in the DCT domain enhances the model's capacity to extract high-frequency information. The variant without both WP and DCT (denoted as W \& D) yields the poorest performance, further confirming that each component independently contributes to the overall improvement. Figure \ref{fig:abhigh} presents visual ablation results on the high-frequency components, where our method produces the best details and textures.  

\noindent \textbf{Ablation for FW-KAN}. In Table \ref{tab:kan}, we compare our FW-KAN with other nonlinear operators, specifically MLP and the original KAN. For MLP, we employ a two-layer activation structure, defined as linear-ReLU-linear-ReLU-linear, to enhance nonlinear expressiveness. MLP-6, MLP-12, and MLP-24 refer to MLPs with 6, 12, and 24 layers. Experiments show that, compared to MLP, our FW-KAN can learn complex representations with fewer parameters, while the original KAN encounters optimization challenges.

\noindent \textbf{Ablation for the two-stage variants}. To evaluate the effectiveness of our multi-stage framework, we perform a two-stage ablation study, as shown in Table \ref{tab:two_stage}. Specifically, we remove the zero-frequency enhancement stage in ERR, resulting in a two-stage paradigm. Since HFR is designed to learn high-frequency components, it remains unchanged, and only the first-stage network is modified. We design three two-stage variants: (1) V1 combines ZFE and LFR into a single network for low frequency restoration; (2) V2 retains only ZFE; and (3) V3 retains only LFR. For fair comparison, we adjust the model capacity by increasing the number of blocks. Results in Table \ref{tab:two_stage} highlight the superiority of the full ERR framework.

\noindent \textbf{Analysis of the model complexity}.
To conduct an in-depth analysis of the trade-off between performance gains and model complexity, we construct a series of models by varying the number of blocks and channels based on our methods and the UHDformer. As shown in Figure~\ref{fig:complex}, we provide a detailed comparison on the UHD-Blur and UHD-Haze datasets, demonstrating that our ERR and IERR consistently outperform UHDformer under comparable parameter budgets.

\subsubsection{Ablation study for IERR}

\begin{figure}[!t]
	\centering
	\includegraphics[width=0.98\linewidth]{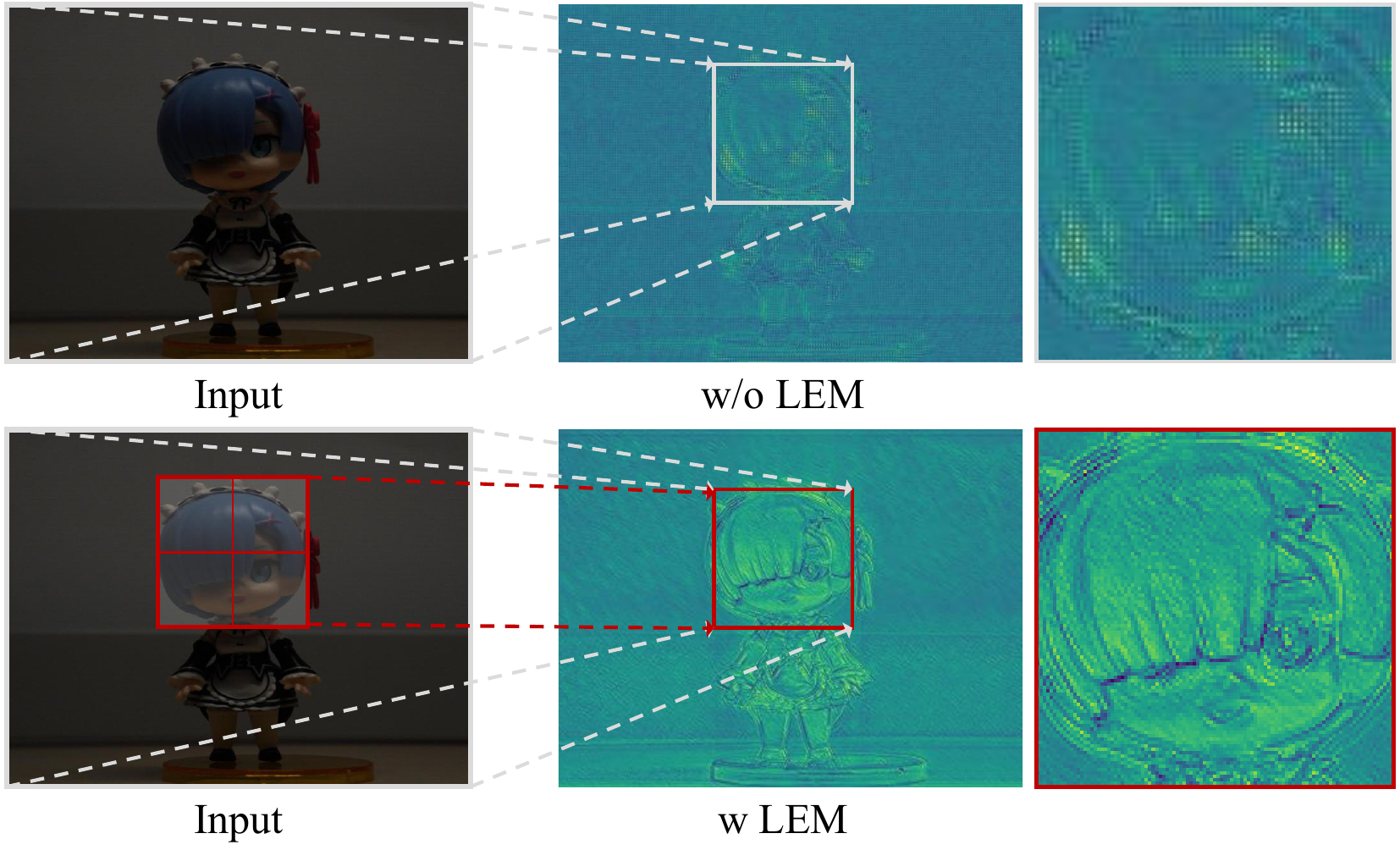}
     \vspace{-0.5em}
	\caption{Analysis of feature visualizations for the designed LEM. The orange dashed line denotes global modeling via SSM, while the red dashed line represents local modeling via our LEM. This comparison demonstrates that incorporating local modeling into the LFR can substantially enhance the feature representation capacity.  } 
	\vspace{-0.1cm}
	\label{fig:LEM}
\end{figure}

In this section, we conduct an ablation study to evaluate the design choices of our extended method, IERR. The improvements in IERR can be attributed to three key components: (1) the enhanced ZFE, which incorporates a global prior projector and operates in a lower-resolution space; (2) the improved LFR, which introduces a local enhancement module; and (3) the  refined HFR, which adopts a zigzag reordering (ZR) strategy.

\noindent \textbf{For ZFE.} As shown in Table~\ref{tab:IERR}, each design choice in ZFE demonstrates effectiveness. Model I  verifies that introducing a prior enhancement network between prior extraction and fusion improves overall performance. Model II further confirms that imposing constraints on the prior information facilitates better learning of global representations. Model III suggests that performing ZFE in an extremely low-resolution space has negligible impact on performance.

\noindent \textbf{For LFR.} To boost the modeling of local features, IERR incorporates a local enhancement module (LEM). It mainly uses convolution operations, and by employing large-kernel convolutions, it effectively widens the receptive field. Model IV with the LEM in Table~\ref{tab:IERR} shows that incorporating the LEM effectively enhances performance. Figure \ref{fig:LEM} illustrates the feature visualization analysis of our LEM, highlighting its significant improvement in local modeling capability.

\noindent \textbf{For HFR.} Table~\ref{tab:IERR}  clearly demonstrates that IERR surpasses Model IV, highlighting the benefit of the zigzag reordering (ZR) strategy. Unlike the conference version that partitions the DCT space without frequency awareness, ZR reorganizes the coefficients from low to high frequency, leading to more coherent frequency grouping and  improved frequency-domain learning.

\begin{table}[t]
    \caption{Comparison of efficiency on 4K images.}
        \vspace{-0.8em}
	\centering

	\resizebox{1\linewidth}{!}{
  {\setlength{\tabcolsep}{3pt}
		\begin{tabular}{c c c c c  }
			\specialrule{1.2pt}{0.2pt}{1pt}
            Model& Param(M)$\downarrow$&FLOPs$\downarrow$&Memory(MiB)$\downarrow$&latency(s)$\downarrow$ \\
			\midrule
			UHDformer  &    0.339   &    385.47     &   15416  &       0.88  \\
            WaveMamba  &    1.258  &   948.75     &  16646  &       0.61   \\
            UDR-mixer  &    4.90   &   1557.89     &  14770  &   0.68       \\
            D2Net  &   5.22    &  4744.54      & 21372   &   2.04       \\
            \midrule
            ERR  &    1.131   &   307.52     &  9880  &       0.52   \\
            IERR  &    0.503  &   288.81     &  9771   &       0.33   \\

			\specialrule{1.2pt}{0.2pt}{1pt}
	\end{tabular}}
    }
    \vspace{-1 em}
	\label{tab:time}
	
\end{table}

\noindent \textbf{Inference efficiency}. Table \ref{tab:time} compares the computational efficiency of our method with recent UHD restoration algorithms. Compared to previous approaches, our IERR achieves the best performance in terms of FLOPs, GPU memory consumption, and inference speed, while also achieving an impressively small parameter count. Notably, both ERR and IERR enable full-resolution 4K image inference on the GTX 1080Ti GPU (11GB memory), which is beyond the capability of previous methods. 


\section{Conclusion}
In this paper, we propose a novel framework named ERR, which consists of three dedicated sub-networks: the Zero-Frequency Enhancer (ZFE) for capturing global contextual information, the Low-Frequency Restorer (LFR) for reconstructing the primary image content, and the High-Frequency Refiner (HFR) equipped with a Frequency-Windowed KAN (FW-KAN) for detail enhancement. 
Additionally, we construct a large-scale, high-quality UHD image dataset to support research in UHD image restoration. 

\noindent \textbf{Acknowledgments.} This work was supported by Natural Science Foundation of China: No. 62406135, Natural Science Foundation of Jiangsu Province: BK20241198, Gusu Innovation and Entrepreneur Leading Talents: No. ZXL2024362.

\ifCLASSOPTIONcaptionsoff
  \newpage
\fi

\small{
\bibliographystyle{IEEEtran}
\bibliography{referencex}
}

\newpage

\end{document}